%% file: main.tex
\documentclass[10pt,journal,compsoc]{IEEEtran}
\usepackage{hyperref}
\usepackage{graphicx}
\usepackage{amsmath}
\usepackage{amssymb}
\usepackage{booktabs}

\usepackage[utf8]{inputenc} 
\usepackage[T1]{fontenc}    
\usepackage{url}            
\usepackage{booktabs}       
\usepackage{amsfonts}       
\usepackage{nicefrac}       
\usepackage{microtype}      
\usepackage{xcolor}         
\usepackage{soul}

\usepackage{unicode}

\usepackage{epsfig}
\usepackage{graphicx}
\usepackage{amsmath}
\usepackage{amssymb}
\usepackage{multirow}
\usepackage{xcolor}
\usepackage{booktabs}
\usepackage{url}

\usepackage{colortbl}
\usepackage{bm}
\usepackage{subcaption}
\usepackage{makecell}
\usepackage{pifont}
\usepackage{enumitem}
\usepackage{wrapfig}
\usepackage[ruled]{algorithm2e}

\definecolor{gain}{rgb}{0.224, 0.710, 0.290}

\newcommand{\del}[1]{}

\newcommand{\etal}[1]{\textit{et al.}}

\newtheorem{prop}{Proposition}

\ifCLASSOPTIONcompsoc
  \usepackage[nocompress]{cite}
\else
  \usepackage{cite}
\fi
\ifCLASSINFOpdf
\else
\fi
\begin{document}
%
\title{SP$^2$OT: Semantic-Regularized Progressive Partial Optimal Transport for Imbalanced Clustering}
%
%
%
%

\author{Chuyu~Zhang,
        Hui~Ren,
        and~Xuming~He
\IEEEcompsocitemizethanks{
\IEEEcompsocthanksitem Chuyu~Zhang is with the ShanghaiTech University and Lingang Laboratory, Shanghai, 201210, China.
E-mail: zhangchy2@shanghaitech.edu.cn
\IEEEcompsocthanksitem Hui~Ren is with the ShanghaiTech University, Shanghai 201210, China.
E-mail: renhui@shanghaitech.edu.cn
\IEEEcompsocthanksitem Xuming~He is with the ShanghaiTech University and Shanghai Engineering Research Center of Intelligent Vision and Imaging, Pudong, Shanghai 201210, China.
E-mail: hexm@shanghaitech.edu.cn
}
\thanks{Manuscript received April 19, 2024; revised August 26, 2015.}}

%
%

\markboth{Journal of \LaTeX\ Class Files,~Vol.~14, No.~8, August~2015}%
{Shell \MakeLowercase{\textit{et al.}}: Bare Demo of IEEEtran.cls for Computer Society Journals}
%



\IEEEtitleabstractindextext{%
\begin{abstract}
Deep clustering, which learns representation and semantic clustering without labels information, poses a great challenge for deep learning-based approaches. Despite significant progress in recent years, most existing methods focus on uniformly distributed datasets, significantly limiting the practical applicability of their methods. In this paper, we propose a more practical problem setting named deep imbalanced clustering, where the underlying classes exhibit an imbalance distribution. To address this challenge, we introduce a novel optimal transport-based pseudo-label learning framework. Our framework formulates pseudo-label generation as a Semantic-regularized Progressive Partial Optimal Transport (SP$^2$OT) problem, which progressively transports each sample to imbalanced clusters under 
prior 
and semantic relation constraints, thus generating high-quality and imbalance-aware pseudo-labels. To solve the SP$^2$OT problem, we propose a projected mirror descent algorithm, which alternates between: (1) computing the gradient of the SP$^2$OT objective, and (2) performing gradient descent with projection via an entropy-regularized progressive partial optimal transport formulation. Furthermore, we formulate the second step as an unbalanced optimal transport problem with augmented constraints and develop an efficient solution based on fast matrix scaling algorithms.
Experiments on various datasets, including a human-curated long-tailed CIFAR100, challenging ImageNet-R, and large-scale subsets of fine-grained iNaturalist2018 datasets, demonstrate the superiority of our method. Code is available: \href{https://github.com/rhfeiyang/SPPOT}{https://github.com/rhfeiyang/SPPOT}.



\end{abstract}

}

\maketitle
\IEEEdisplaynontitleabstractindextext

%
\IEEEpeerreviewmaketitle

\input{sec/intro.tex}
\input{sec/related}
\input{sec/method}

\input{sec/exp}
\input{sec/conclu}
\ifCLASSOPTIONcaptionsoff
  \newpage
\fi



%
{	\small
	\bibliographystyle{IEEEtran}
	\bibliography{main}
}

%




\begin{IEEEbiography}[{\includegraphics[width=1in,height=1.25in,clip,keepaspectratio]{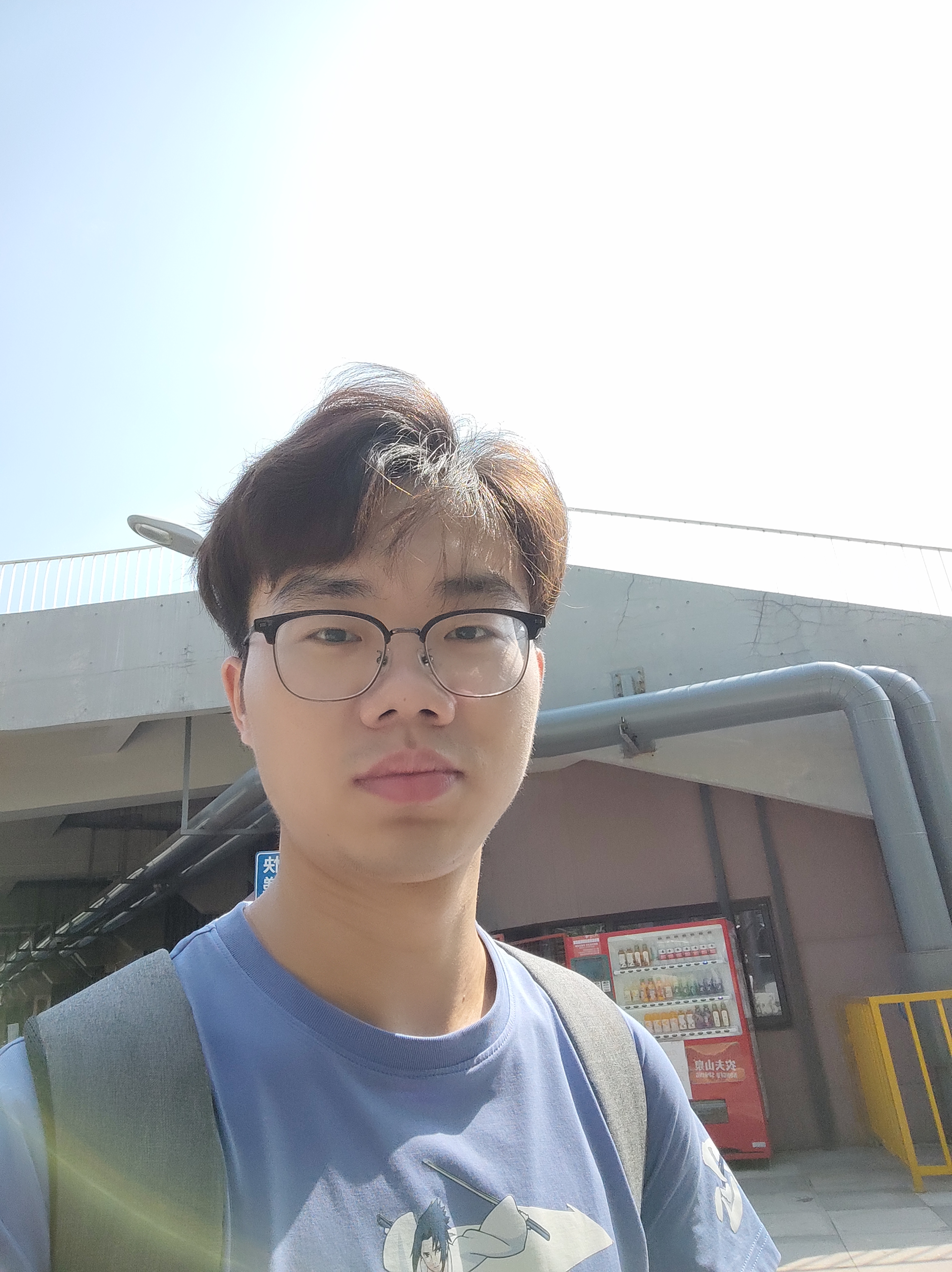}}]{Chuyu Zhang}
received the B.E. degree in the School of Electronic Information Engineering from Wuhan University, Wuhan, China, in 2020. He is currently pursuing the Ph.D. degree in computer science and technology at ShanghaiTech University, supervised by Prof. Xuming He. His research interests concern computer vision and machine learning.
\end{IEEEbiography}

\begin{IEEEbiography}[{\includegraphics[width=1in,height=1.25in,clip,keepaspectratio]{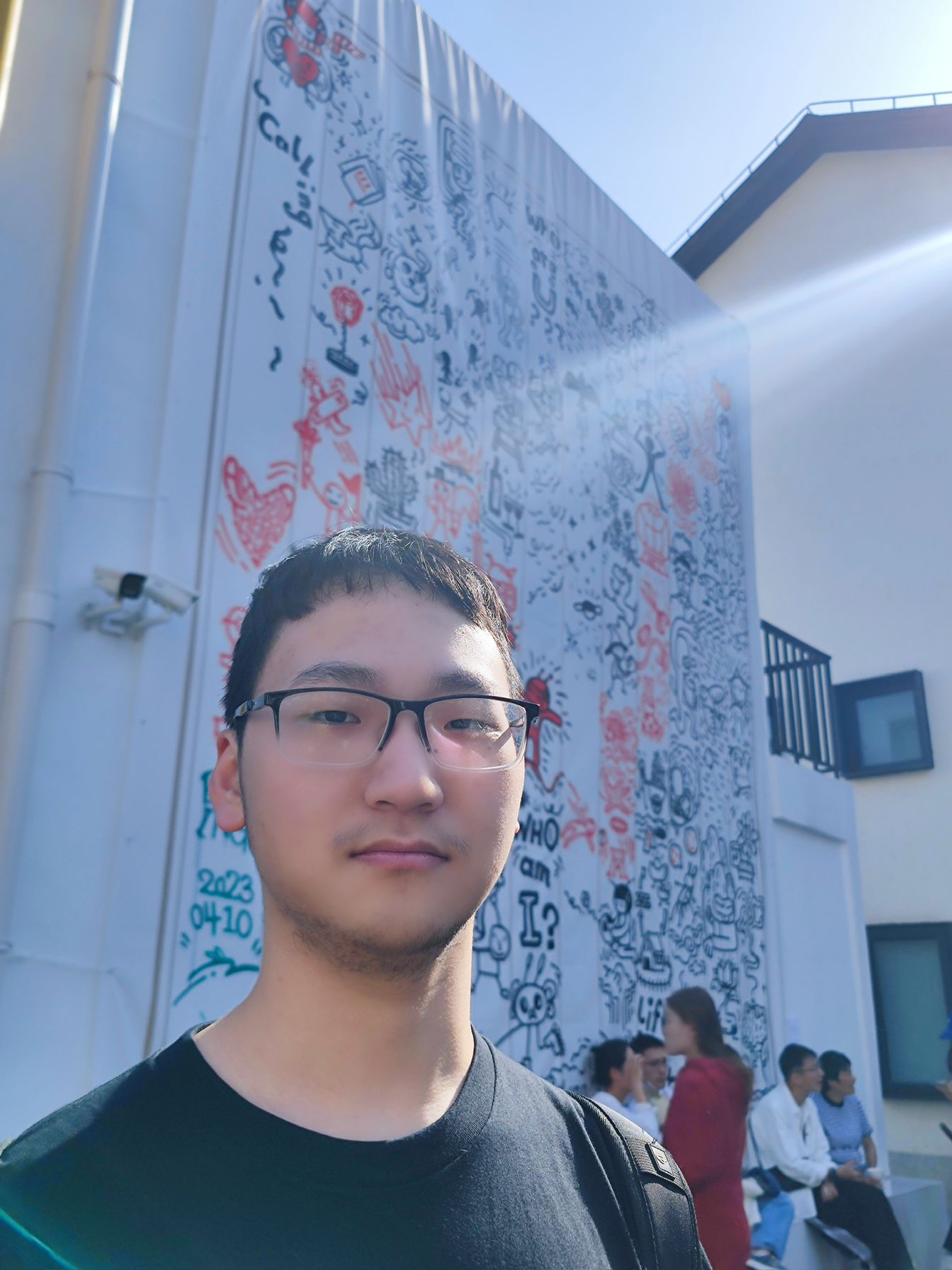}}]{Hui Ren}
is currently pursuing a B.E. degree in computer science and technology at ShanghaiTech University. He is an undergraduate research assistant in PLUS Lab at ShanghaiTech University, advised by Prof. Xuming He. His research interests involve machine learning and computer vision. Personal Website: https://rhfeiyang.github.io
\end{IEEEbiography}



\begin{IEEEbiography}[{\includegraphics[width=1in,height=1.25in,clip,keepaspectratio]{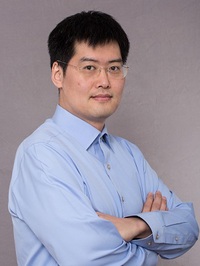}}]{Xuming He}
received the Ph.D. degree in computer science from the University of Toronto, Toronto, ON, Canada, in 2008. He held a Post-Doctoral position with the University of California at Los Angeles, Los Angeles, CA, USA, from 2008 to 2010. He was an Adjunct Research Fellow with The Australian National University, Canberra, ACT, Australia, from 2010 to 2016. He joined National ICT Australia, Canberra, where he was a Senior Researcher from 2013 to 2016. He is currently an Associate Professor with the School of Information Science and Technology, ShanghaiTech University, Shanghai, China. His research interests include semantic image and video segmentation, 3-D scene understanding, visual motion analysis, and efficient inference and learning in structured models.
\end{IEEEbiography}

\end{document}


%
\title{SP$^2$OT: Semantic-Regularized Progressive Partial Optimal Transport for Imbalanced Clustering}
%
%
%
%

\author{Chuyu~Zhang,
        Hui~Ren,
        and~Xuming~He
\IEEEcompsocitemizethanks{
\IEEEcompsocthanksitem Chuyu~Zhang is with the ShanghaiTech University and Lingang Laboratory, Shanghai, 200031, China.
E-mail: zhangchy2@shanghaitech.edu.cn
\IEEEcompsocthanksitem Hui~Ren is with the ShanghaiTech University, Shanghai 201210, China.
E-mail: renhui@shanghaitech.edu.cn
\IEEEcompsocthanksitem Xuming~He is with the ShanghaiTech University and Shanghai Engineering Research Center of Intelligent Vision and Imaging, Pudong, Shanghai 201210, China.
E-mail: hexm@shanghaitech.edu.cn
}
\thanks{Manuscript received April 19, 2024; revised August 26, 2015.}}

%
%

\markboth{Journal of \LaTeX\ Class Files,~Vol.~14, No.~8, August~2015}%
{Shell \MakeLowercase{\textit{et al.}}: Bare Demo of IEEEtran.cls for Computer Society Journals}
\maketitle
\IEEEdisplaynontitleabstractindextext

%
\IEEEpeerreviewmaketitle

\input{sec/supply}

\ifCLASSOPTIONcaptionsoff
  \newpage
\fi



%
{	\small
	\bibliographystyle{IEEEtran}
	\bibliography{main}
}

%
\begin{IEEEbiography}[{\includegraphics[width=1in,height=1.25in,clip,keepaspectratio]{./imgs/photos/zcy.jpg}}]{Chuyu Zhang}
received the B.E. degree in the School of Electronic Information Engineering from Wuhan University, Wuhan, China, in 2019. He is currently pursuing the Ph.D. degree in computer science and technology at ShanghaiTech University, supervised by Prof. Xuming He. His research interests concern computer vision and machine learning.
\end{IEEEbiography}

\begin{IEEEbiography}[{\includegraphics[width=1in,height=1.25in,clip,keepaspectratio]{./imgs/photos/rh.jpg}}]{Hui Ren}
is currently pursuing a B.E. degree in computer science and technology at ShanghaiTech University. He is an undergraduate research assistant in PLUS Lab at ShanghaiTech University, advised by Prof. Xuming He. His research interests involve machine learning and computer vision. Personal Website: https://rhfeiyang.github.io
\end{IEEEbiography}


\begin{IEEEbiography}[{\includegraphics[width=1in,height=1.25in,clip,keepaspectratio]{./imgs/photos/hexm.jpeg}}]{Xuming He}
received the Ph.D. degree in computer science from the University of Toronto, Toronto, ON, Canada, in 2008. He held a Post-Doctoral position with the University of California at Los Angeles, Los Angeles, CA, USA, from 2008 to 2010. He was an Adjunct Research Fellow with The Australian National University, Canberra, ACT, Australia, from 2010 to 2016. He joined National ICT Australia, Canberra, where he was a Senior Researcher from 2013 to 2016. He is currently an Associate Professor with the School of Information Science and Technology, ShanghaiTech University, Shanghai, China. His research interests include semantic image and video segmentation, 3-D scene understanding, visual motion analysis, and efficient inference and learning in structured models.
\end{IEEEbiography}


%% file: sec/intro.tex
\section{Introduction}\label{sec:intro}


\IEEEPARstart{H}umans possess an inherent ability to categorize similar concepts, even when encountering them for the first time. However, artificial models face challenges in grouping similar concepts without explicit labels, highlighting the limitations of traditional supervised learning approaches. Recognizing this gap, the field of deep clustering has emerged as a promising avenue. Deep clustering goes beyond supervised learning approaches by simultaneously learning effective representations and semantic clustering, enabling models to discern and group similar concepts in an unsupervised manner. Several seminal works, including \cite{ji2019invariant, vangansbeke2020scan, Ronen:CVPR:2022:DeepDPM}, have made significant contributions to advancing the field of deep clustering. However, a prevalent limitation in existing research lies in its emphasis on developing methods tailored for balanced datasets. While these approaches showcase promising results in controlled environments, their practical applicability is restricted due to prevalent imbalanced data distributions in real-world scenarios. Consequently, there is a growing need to extend the capabilities of deep clustering methods to handle imbalanced datasets effectively, ensuring their robust performance across a broader spectrum of real-world applications. 

In this paper, we consider a practical deep imbalanced clustering problem, bridging the gap between existing deep clustering methodologies and realistic settings. Current approaches can be broadly classified into three categories: relation matching-based~\cite{Dang_2021_CVPR}, mutual information maximization-based~\cite{ji2019invariant}, and pseudo-labeling-based~\cite{vangansbeke2020scan} methods. They face significant challenges when dealing with imbalanced clustering. 
Specifically, relation matching and mutual information maximization-based methods adopt surrogate losses for clustering and disregard imbalanced cluster distributions, potentially leading to subpar outcomes in imbalanced scenarios~\cite{jiang2021self,zhou2022contrastive}. 
Pseudo-labeling suffers from three drawbacks in deep imbalanced clustering.
Firstly, it heavily depends on the assumption of a uniform distribution, which causes it to fail in handling imbalanced scenarios, an issue explored in Sec. \ref{main_results}. Secondly, it usually necessitates an additional phase for initializing representations~\cite{vangansbeke2020scan,niu2022spice} and requires tedious hyperparameter calibration to alleviate confirmation bias~\cite{arazo2020pseudo}.
Lastly, it primarily generates pseudo-labels by utilizing information in the label space without any adjustment. Consequently, this approach is prone to bias in imbalanced scenarios~\cite{kang2019decoupling} and fails to leverage the rich semantic information inherent in the representation space.

To mitigate the aforementioned weaknesses, we introduce a novel progressive learning framework based on pseudo-labeling. This framework formulates the pseudo-label generation as a Semantic-regularized Progressive Partial Optimal Transport (SP$^2$OT) algorithm. This SP$^2$OT seamlessly integrates three critical components into a unified optimal transport (OT) problem: K-Nearest-Neighbor (KNN) graph-based semantic relation regularization, modeling of imbalanced class distributions, and the selection of confident samples. This novel integration empowers our algorithm to generate high-quality and imbalanced pseudo-labels by considering both the output and semantic space simultaneously, while alleviating confirmation bias through progressive learning from high-confidence samples.
In addition, we propose an efficient solver for the SP$^2$OT problem based on the projected mirror descent algorithm.



Specifically, our framework performs clustering by generating pseudo-labels and learning from them alternately. We formulate the pseudo-label generation as the SP$^2$OT algorithm, which assigns each sample to clusters with minimal cost. The transportation cost is defined as the negative log of model predictions, and the assignment is subject to constraints on sample relations in the representation space and a set of prior constraints. The key constraints in our formulation are: 1) Distribution Constraint: We employ a Kullback-Leibler (KL) divergence constraint to ensure a uniform distribution of cluster sizes. This constraint is crucial for preventing degenerate solutions while allowing for the generation of imbalanced pseudo-labels due to its relaxed nature compared to an equality constraint. 2) Total Mass Constraint: By adjusting sample weights through optimization, the total mass constraint enables selective learning from high-confidence samples. This mitigates the impact of noisy pseudo-labels and eliminates the need for sensitive hand-crafted confidence threshold tuning. Moreover, incrementally raising the total mass constraint facilitates a gradual transition from learning easy samples to tackling more challenging ones. 3) Semantic Relation Regularization: We incorporate a semantic relation regularization term represented by a K-Nearest Neighbor (KNN) graph. This regularization ensures that similar pseudo-labels are generated for the nearest samples in the representation space, enhancing the overall quality of the pseudo-labels.

We solve the SP$^2$OT problem using projected mirror descent (PMD)~\cite{peyre2016gromov}, which alternates between two key steps: (1) gradient computation of the SP$^2$OT objective and (2) gradient descent with projection via an entropy-regularized progressive partial optimal transport (P$^2$OT) formulation~\cite{zhang2024p2ot}. To improve computational efficiency, we propose two key enhancements to P$^2$OT: (1) introducing a virtual cluster structure and (2) replacing the standard KL divergence with a weighted KL constraint. This reformulation casts P$^2$OT as an unbalanced optimal transport problem that can be efficiently solved using a matrix-scaling algorithm. Our experimental results demonstrate significant performance improvements, with speed gains reaching 6.2$\times$ compared to traditional generalized scaling approaches.


Compared to the conference version~\cite{zhang2024p2ot}, which generates pseudo-labels using only the output space, this work introduces two important advancements: 1) we propose SP$^2$OT, which enable P$^2$OT to effectively leverage semantic information; 2) building upon the PMD algorithm, we extend the solver from \cite{zhang2024p2ot} to handle unbalanced fused Gromov-Wasserstein optimal transport problems, significantly broadening the applicability of the original framework.



To validate our method, we conduct experiments on a wide range of challenging datasets, including the human-curated CIFAR100 dataset~\cite{krizhevsky2009learning}, challenging `out-of-distribution' ImageNet-R dataset~\cite{hendrycks2021many}, and large-scale subsets of fine-grained iNaturalist18 dataset~\cite{van2018inaturalist}. Experiments on those challenging datasets demonstrate the effectiveness of each component and the superiority of our method.

In summary, our contribution is as follows:

\begin{itemize}
    \item We generalize the deep clustering problem to more realistic imbalance scenarios, and establish a new benchmark. 
    \item We introduce a pseudo-label learning framework for deep imbalance clustering, along with a novel SP$^2$OT algorithm. This approach enables us to effectively account for semantic relations, class imbalance distribution, and progressive learning during pseudo-label generation.
    \item We propose an efficient SP$^2$OT solver based on projected mirror descent, alternating between gradient computation and gradient descent with projection steps—where projection is reformulated as unbalanced optimal transport problems and solved via fast matrix-scaling algorithms.
    \item Our method achieves the SOTA performance on most datasets compared with existing methods on our newly proposed challenging and large-scale benchmark.
\end{itemize}

%% file: sec/related.tex
\section{Related Work}

\subsection{Deep clustering} 
The goal of deep clustering~\cite{zhou2022comprehensive,huang2022learning,li2023image} is to learn representation and cluster data into semantic classes simultaneously. Based on clustering methods, current works can be grouped into three categories: relation matching, mutual information maximization, and pseudo labeling. Specifically, relation matching~\cite{chang2017deep,vangansbeke2020scan,tao2020clustering} involves the minimization of the distance between instances considered `similar' by representation distance, thus achieving clustering in a bottom-up way. Mutual information maximization~\cite{ji2019invariant,li2021contrastive,shen2021you} either in the prediction or representation space, aims at learning invariant representations for images with different augmentation, thereby achieving clustering implicitly~\cite{ben2023reverse}. Pseudo labeling (PL)~\cite{lee2013pseudo}, which is the focus of this paper and widely used in semi-supervised learning ~\cite{sohn2020fixmatch,zhang2021flexmatch}, assigns each instance to semantic clusters and has demonstrated exceptional performance in the domain of deep clustering~\cite{caron2018deep,vangansbeke2020scan,asano2020self,niu2022spice}. 
In contrast to the relatively clean pseudo-labels produced by semi-supervised learning, deep clustering often generates noisier pseudo-labels, particularly in scenarios characterized by data imbalance, imposing a great challenge on pseudo-label generation. 

To acquire high-quality pseudo labels, \cite{vangansbeke2020scan,niu2022spice} often need multiple-stage training and meticulous manual selection of confidence samples. Specifically, Van \etal ~\cite{vangansbeke2020scan} initially focuses on learning clustering by minimizing the distance between similar images. Subsequently, they undertake confidence sample selection employing hand-crafted thresholding, with variations tailored to different datasets. Based on the unrealistic uniform assumption, Niu \etal ~\cite{niu2022spice} select an equal number of reliable samples for each class based on the confidence ratio and apply FixMatch-like training in the third training stage, which usually fails in the imbalance scenarios.

By contrast, instead of relying on the strong uniform distribution constraint and manual sample selection, our PL-based SP$^2$OT algorithm progressively generates high-quality pseudo labels by simultaneously considering class imbalance distribution, sample confidence, and semantic regularization through a single convex optimization problem.



\vspace{-0.5em}
\subsection{Self-supervised Learning}
Self-supervised Learning (SSL)~\cite{bengio2013representation,jing2020self} aims to learn meaningful representation from vast unlabeled data, and implicitly performs clustering~\cite{weng2025clustering}. In our perspective, SSL methodologies can be broadly categorized into three groups: contrastive learning~\cite{chen2020simple,he2020momentum,wang2022contrastive,xu2022seed}, self-distillation~\cite{caron2021emerging,chen2021exploring,grill2020bootstrap}, and reconstruction-based~\cite{huang2023contrastive,he2022masked,radosavovic2023real} approaches. Contrastive learning methods operate on the principle of encouraging similarity between semantically transformed versions of an input. Self-distillation methods focus on learning invariant predictions from different views while avoiding collapsing through variant techniques. Reconstruction-based methods, on the other hand, aim to predict the original image through generative models or masked image modeling. For a more comprehensive exploration of these SSL methods, we direct the reader to~\cite{balestriero2023cookbook}. It is noteworthy that these SSL techniques can be used as a robust foundational model for deep clustering.

In SSL, the works mostly related to us are \cite{jiang2021self,zhou2022contrastive,liu2021self,yigeometry,hou2023subclass}, which aim to learn representation from imbalanced data. To defend the bias introduced by imbalanced data, Jiang \etal ~\cite{jiang2021self} propose self-damaging contrastive learning to emphasize the most easily forgotten samples, Liu \etal ~\cite{liu2021self} devise a re-weighted regularization technique based on theoretical insights, and Zhou \etal ~\cite{zhou2022contrastive} leverages
the memorization effect of deep neural networks
to automatically drive the information discrepancy of the sample views in contrastive learning. ~\cite{yigeometry} proposes a long-tailed representation learning method to arrange features from different classes into symmetrical and linearly separable regions. In contrast to these methodologies, our approach involves simultaneous representation learning through pseudo-labeling and the acquisition of cluster heads from imbalanced data.
\vspace{-0.5em}
\subsection{Supervised Long-tailed Learning}
Supervised long-tailed learning~\cite{cao2019learning,Zhang_2021_CVPR} is designed to acquire unbiased representations and classifiers from datasets with imbalanced class distributions. Various methods have been proposed to enhance the learning of tail classes by leveraging information about the known class distribution. For instance, Logit Adjustment~\cite{menon2020long} addresses classifier bias by adjusting the logits based on class frequencies. Kang et al.~\cite{kang2019decoupling} take a different approach by decoupling representation and classifier learning. They first learn representations through instance-balanced resampling and then retrain classifiers using re-balancing techniques. For a more extensive survey, readers are encouraged to refer to \cite{zhang2023deep}.  In contrast to supervised long-tailed learning, deep imbalanced clustering faces the challenge of dealing with an unknown and imbalanced distribution, rendering the aforementioned techniques challenging to apply.
\vspace{-0.5em}
\subsection{Optimal transport and its application} 
Optimal Transport (OT)~\cite{villani2009optimal,peyre2017computational,khamis2023earth,lahn2024combinatorial,phatak2022computing} aims to find the most efficient transportation plan while adhering to marginal distribution constraints. It has been used in a broad spectrum of machine learning tasks, including but not limited to generative model~\cite{frogner2015learning,gulrajani2017improved}, semi-supervised learning~\cite{tai2021sinkhorn,taherkhani2020transporting}, clustering~\cite{asano2020self,caron2020unsupervised}, domain adaptation~\cite{flamary2016optimal,chang2022unified,Liu_2023_CVPR,chang2023csot} and others~\cite{wang2022optimal,luo2022differentiable}. Of particular relevance to our work is its utilization in pseudo labeling~\cite{asano2020self,tai2021sinkhorn,zhang2023novel}. Compared to naive pseudo labeling ~\cite{lee2013pseudo}, OT can consider class distribution to generate high-quality pseudo labels and avoid the degenerate solution. Specifically, Asano \etal ~\cite{asano2020self} initially introduces an equality constraint on cluster size, formulating pseudo-label generation as an optimal transport problem. Subsequently, Tai \etal ~\cite{tai2021sinkhorn} enhances the flexibility of optimal transport by incorporating relaxation on the constraint, which results in its failure in deep imbalanced clustering, and introduces label annealing strategies through an additional total mass constraint. Recently, Zhang \etal ~\cite{zhang2023novel} relaxes the equality constraint to a $KL$ divergence constraint on cluster size, thereby addressing imbalanced data scenarios. In contrast to these approaches, which either ignore class imbalance distribution or the confidence of samples, our SP$^2$OT algorithm takes both into account simultaneously, allowing us to generate pseudo-labels progressively and with an awareness of class imbalance.

The Gromov-Wasserstein (GW) distance~\cite{vayer2020fused,bai2025partial,memoli2011gromov}, designed for comparing non-aligned spaces, shares conceptual similarities with our formulation—particularly the fused Gromov-Wasserstein optimal transport (FGWOT)~\cite{memoli2011gromov,peyre2016gromov,xu2024temporally}. While the GW distance measures discrepancies between metric measure spaces ~\cite{peyre2016gromov}, our formulation introduces an additional constraint to enforce structural similarity across datasets in the two spaces. Notably, our approach can be derived as a simplified variant of FGWOT under stronger assumptions (Refer to Sec.\ref{sec:semantic_constraint} for more discussion). 
Furthermore, unlike \cite{xu2024temporally}, we incorporate a total mass constraint to regulate pseudo-label quality and propose an efficient scaling algorithm for faster computation.

%% file: sec/method.tex
\vspace{-0.5em}
\section{Preliminary}
\subsection{Optimal Transport for Pseudo Labeling}

Optimal Transport (OT) tackles the general problem of moving one distribution of mass to another with minimal cost. Mathematically, given two probability vectors $\bm\mu\in\mathbb{R}^{m\times 1},\bm\nu\in\mathbb{R}^{n\times 1}$, as well as a cost matrix $\mathbf{C} \in \mathbb{R}^{m\times n}$ defined on joint space, the objective function which OT minimizes is as follows:
\begin{equation}\label{eq:general_form}
\min_{\mathbf Q \in \mathbb{R}^{m\times n}}\langle\mathbf{Q},\mathbf C\rangle_F + F_1(\mathbf{Q}\mathbf{1}_{n}, \bm\mu) + F_2(\mathbf{Q}^{\top} \mathbf1_m, \bm\nu)
\end{equation}
where $\mathbf Q \in \mathbb{R}^{m\times n}$ is the transportation plan, $\langle,\rangle_F$ denotes the Frobenius product, $F_1,F_2$ are constraints on the marginal distribution of $\mathbf Q$ and $\bm 1_n \in \mathbb{R}^{n\times 1}, \bm 1_m \in \mathbb{R}^{m\times 1}$ are all ones vector. For example, if $F_1, F_2$ are equality constraints, i.e. $\mathbf{Q}\mathbf{1}_{n}=\bm\mu, \mathbf{Q}^{\top} \mathbf1_m = \bm\nu$, the above OT becomes a widely-known Kantorovich's form~\cite{kantorovich}. And if $F_1, F_2$ are $KL$ divergence or inequality constraints, Eq.(\ref{eq:general_form}) turns into the unbalanced OT problem~\cite{liero2018optimal}. 

In practice, to efficiently solve Kantorovich's form OT problem, Cuturi~\cite{cuturi2013sinkhorn} introduces an entropy term $-\epsilon\mathcal{H}(\mathbf{Q})$ to Eq.(\ref{eq:general_form}) and solve the entropic regularized OT with the efficient scaling algorithm~\cite{knight2008sinkhorn}. Subsequently, Chizat \etal ~\cite{chizat2018scaling} generalizes the scaling algorithm to solve the unbalanced OT problem. Therefore, Eq.(\ref{eq:general_form}) can be solved by Algorithm \ref{alg:saot}. We show more detail of the scaling algorithm in the Appendix 1.
Additionally, one can introduce the total mass constraint to Eq.(\ref{eq:general_form}), and it can be solved by the efficient scaling algorithm by adding a dummy or virtual point to absorb the total mass constraint into marginal. 

Recently, Asano \etal ~\cite{asano2020self} applied optimal transport to generate pseudo labels for unsupervised clustering. Specifically, given the model's prediction and its pseudo-label, the loss function is denoted as follows:

\begin{equation}\label{eq:loss}
    \mathcal{L} = -\sum_{i=1}^{N} \mathbf{Q}_i\log \mathbf{P}_i = \langle \mathbf{Q}, -\log \mathbf P \rangle_F,
\end{equation}
where $\langle,\rangle _F$ is the Frobenius product, $\mathbf Q, \mathbf P \in \mathbb{R}^{N\times K}_+$, $\mathbf Q_i, \mathbf P_i$ is the pseudo label and prediction of sample $x_i$. Note that we have absorbed the normalize term $\frac{1}{N}$ into $\mathbf{Q}$ for simplicity, thus $\mathbf Q\bm 1_K = \frac{1}{N}\bm1_N$.

By making the assumption of equal distribution within the cluster, Asano \etal ~\cite{asano2020self} approach the problem of pseudo-label generation by formulating it as an optimal transport problem, as expressed by the following formulation:
\begin{align}\label{eq:balanced_ot}
\min_{\mathbf{Q} \in \Pi}\langle\mathbf{Q},-\log& \mathbf P\rangle_F \\
\text{s.t.}\ \Pi = \{\mathbf{Q} \in \mathbb{R}^{N\times K}_+ | \mathbf{Q} \bm1_K&=\frac{1}{N}\bm1_N, \mathbf{Q}^\top \bm1_N=\frac{1}{K}\bm 1_K\}
\end{align}
where $\lambda$ is a scalar factor. Intuitively, $\Pi$ in Eq.(4) transports each data point to uniformly distributed clusters with minimal cost based on the distance between the sample and cluster. The authors then employ the Algorithm \ref{alg:saot}, to obtain the optimal $\mathbf{Q}^\star$. Subsequently, utilizing $\mathbf{Q}^\star$, the model is updated using Eq.(\ref{eq:loss}). Interestingly, the clustering is achieved through iterative optimization of Eq.(\ref{eq:balanced_ot}) and Eq.(\ref{eq:loss}), both sharing the same objective. Our algorithm builds upon this formulation, deviating from the balanced cluster assumption and incorporating additional semantic penalty. This modification enhances the algorithm's performance and makes it applicable to more realistic scenarios where the cluster is imbalanced.

\SetKwComment{Comment}{//}{}
\begin{algorithm}[t]
\caption{Scaling Algorithm for OT}\label{alg:saot}
\SetAlgoLined
\footnotesize
\DontPrintSemicolon
\KwIn{Cost matrix $\mathbf C$, $\epsilon,m,n,\bm\mu,\bm\nu$}    

$\mathbf{M}= \exp(-\mathbf{C}/\epsilon)$

$\mathbf{b} \leftarrow \mathbf{1}_{n} $

\While{$\bm b$ not converge}{

$\mathbf{a} \leftarrow \text{prox}_{F_1/\epsilon}^{KL}(\mathbf{M}  \mathbf{b}, \bm\mu)/(\mathbf{M}  \mathbf{b})$

$\mathbf{b} \leftarrow \text{prox}_{F_2/\epsilon}^{KL}(\mathbf{M}^\top  \mathbf{a}, \bm\nu)/(\mathbf{M}^\top  \mathbf{a})$
}

\KwRet $\text{diag}(\mathbf{a}) \mathbf{M} \text{diag}(\mathbf{b})$;

\end{algorithm}

\subsection{Fused Gromov-Wasserstein Optimal Transport}\label{sec:fgwot}
The Gromov-Wasserstein (GW) distance, proposed by \cite{memoli2011gromov}, quantifies the discrepancy between probability distributions supported on different metric spaces by aligning their intrinsic structures. Given two finite metric measure spaces, $\mathcal{Z} = ({z_1, z_2, \ldots, z_{N_1}}, C, p)$ and $\mathcal{Y} = ({y_1, y_2, \ldots, y_{N_2}}, \hat{C}, q)$, the GW distance \cite{peyre2016gromov} between them is defined as:
\begin{equation}\label{eq:gw}
\inf_{\mathbf{Q} \in \Pi(p, q)} \sum_{i,j,k,l} \mathcal{L}(C_{i,k}, \hat{C}_{j,l}) \mathbf{Q}_{i,j} \mathbf{Q}_{k,l}
\end{equation}
where $\mathbf{Q} \in \mathbb{R}^{m \times n}$ is the transportation plan, and $\Pi(p, q)$ denotes the set of all admissible couplings between $p$ and $q$. The loss function $\mathcal{L}(C_{i,k}, \hat{C}_{j,l})$ measures the discrepancy between distances in the two spaces. A common choice is:

\begin{equation}
\mathcal{L}(C_{i,k}, \hat{C}_{j,l}) = \left| C(z_i, z_k) - \hat{C}(y_j, y_l) \right|
\end{equation}

Intuitively, in the GW distance (Eq.(\ref{eq:gw})), the transport plan $\mathbf{Q}$ is computed by aligning the internal distance structures of $\mathcal{Z}$ and $\mathcal{Y}$, whereas in standard optimal transport (Eq.(\ref{eq:general_form})), the transport plan is derived based on the direct point-to-point distances between the two spaces.

When considering both internal distance structures and point-to-point distances simultaneously, we formulate the Fused Gromov-Wasserstein Optimal Transport (F-GWOT) problem~\cite{vayer2020fused} as follows:
\begin{align}\label{eq:gw_general_form}
\min_{\mathbf{Q} \in \mathbb{R}^{m \times n}} \langle \mathbf{Q}, \mathbf{C} \rangle_F + \sum_{i,j,k,l} \mathcal{L}(C_{i,k}, \hat{C}_{j,l}) \mathbf{Q}_{i,j} \mathbf{Q}_{k,l} \notag \\
F_1(\mathbf{Q}\mathbf{1}_n, \bm{\mu}) + F_2(\mathbf{Q}^\top \mathbf{1}_m, \bm{\nu})
\end{align}

To solve this problem efficiently, we employ projected mirror descent~\cite{peyre2016gromov}. This approach computes the gradient, performs gradient descent, and projects the solution onto the feasible set using a single Sinkhorn algorithm, ensuring both scalability and convergence.


\section{Method}

\subsection{Problem Setup and Method Overview}\label{sec:ps}

In deep imbalanced clustering, the training dataset is denoted as $\mathcal{D}=\{{(x_i)}\}^N_{i=1}$, where the cluster labels and distribution are unknown. The number of clusters $K$ is predefined by the user or estimated by other methods. The goal is to learn representation and semantic clusters. To achieve that, we learn representation and generate pseudo-label alternately, improving the data representation and the quality of cluster assignments. Specifically, given the cluster prediction, we utilize our novel Semantic-regularized Progressive Partial Optimal Transport (SP$^2$OT) algorithm to generate high-quality pseudo labels. Our SP$^2$OT algorithm has three advantages: 1) generating pseudo labels in an imbalanced manner; 2) reweighting confident samples through optimization; and 3) incorporating semantic information as a regularization. Then, given the pseudo label, we update the representation. We alternate the above two steps until convergence. 

To provide a comprehensive understanding of our innovative SP$^2$OT algorithm, we begin by deriving the formulation of the P$^2$OT algorithm, which serves as a foundational component without considering the semantic penalty present in SP$^2$OT. Then, we meticulously outline the semantic penalty, leveraging it in conjunction with P$^2$OT to derive the SP$^2$OT algorithm.




\subsection{Progressive Partial Optimal Transport (P$^2$OT)}
In deep clustering, it is typical to impose some constraint on the cluster size distribution to avoid a degenerate solution, where all the samples are assigned to a single cluster. As the cluster distribution is long-tailed and unknown, we adopt a $KL$ divergence constraint and only assume the prior distribution is uniform.
Therefore, with two marginal distribution constraints, we can formulate the pseudo-label generation problem as an unbalanced OT problem:
\begin{align}\label{eq:unbalanced_ot}
\min_{\mathbf{Q} \in \Pi}\langle\mathbf{Q},-\log \mathbf P\rangle_F + \lambda KL(\mathbf{Q}^\top \bm1_N,\frac{1}{K}\bm 1_K) \\
\text{s.t.}\quad \Pi = \{\mathbf{Q} \in \mathbb{R}^{N\times K}_+ | \mathbf{Q} \bm1_K=\frac{1}{N}\bm1_N\}
\end{align}
where $\langle,\rangle _F$ is the Frobenius product, and $\lambda$ is a scalar factor. In Eq.(\ref{eq:unbalanced_ot}), the first term is exactly $\mathcal{L}$, the $KL$ term is a constraint on cluster size, and the equality term ensures each sample is equally important.

However, the unbalanced OT algorithm treats each sample equally, which may generate noisy pseudo labels, due to the initial representation being poor, resulting in confirmation bias. Inspired by curriculum learning~\cite{wang2021survey}, which first learns from easy samples and gradually learns hard samples, we select only a fraction of high-confident samples to learn initially and increase the fraction gradually. However, instead of manually selecting confident samples through sensitive thresholding, we formulate the selection process as a total mass constraint in Eq.(\ref{eq:curr_unbalanced_ot}). This approach allows us to reweight each sample through joint optimization with the pseudo-label generation, eliminating the need for sensitive hyperparameter tuning. Therefore, the formulation of our novel P$^2$OT is as follows:
\begin{align}\label{eq:curr_unbalanced_ot}
\min_{\mathbf{Q} \in \Pi}\langle\mathbf{Q},-\log \mathbf P\rangle_F + \lambda &KL(\mathbf{Q}^\top \bm1_N,\frac{\rho}{K}\bm 1_K) \\
\text{s.t.}\quad \Pi = \{\mathbf{Q} \in \mathbb{R}^{N\times K}_+ | \mathbf{Q} \bm1_K&\leq\frac{1}{N}\bm1_N, \bm 1_N^\top\mathbf{Q} \bm 1_K=\rho\}
\end{align}
where $\rho$ is the fraction of selected mass and will increase gradually, and $KL$ is the unnormalized divergence measure, enabling us to handle imbalance distribution. Under the total mass constraint, the algorithm prioritizes moving samples with minimal cost (higher $\mathbf P$) to each cluster, achieving high-confidence sample selection. The resulting $\mathbf{Q} \bm 1_K$ represents the weights for samples, and we directly utilize it to reweight each sample without any further modification.
Intuitively, P$^2$OT encompasses a set of prior distribution constraints, achieving the generation of imbalanced pseudo-labels through a relaxed $KL$ penalty and reweighting confident samples by total mass constraint within a single optimization problem.


\subsection{Semantic regularization} \label{sec:semantic_constraint}
By now, our algorithm is able to generate pseudo-labels solely based on the model's predictions in the output space. However, it is noteworthy that the model's predictions may lack reliability, particularly during the early stages of training, and it tends to bias to majority classes in imbalanced scenarios~\cite{menon2020long}.
On the other hand, the proposed P$^2$OT only considers the point-wise prediction and statistical distribution constraints to generate pseudo-labels, which ignores the valuable semantic relation among samples, resulting in sub-optimal outcomes.

To address the issues mentioned, we propose incorporating the semantic information obtained from the feature space of the pre-trained model backbone, which is more robust to imbalanced data, into the pseudo-label generation process. Specifically, given the feature representation of all samples in the dataset $\mathbf{z} = \{z_1, ..., z_N\} \in \mathbb{R}^{N\times D}$, we construct a K-Nearest Neighbor Graph (K-NNG) to capture reliable and meaningful semantic relations, such as manifold structures among samples. Where the adjacency matrix $\mathbf{A} \in \mathbb{R}^{N\times N}$ is defined as follows:
\begin{align}\label{eq:adjacency1}
    \mathbf{S}: \mathbf{S}_{i,j} = \mathit{K}(z_i, z_j) &= \exp(- \frac{|| z_i- z_j ||^2}{2\sigma^2} )\\ \label{eq:adjacency2}
    \mathbf{A} &= \mathit{F} ( \mathbf{S} ),
\end{align}
where $\mathit{K}$ is a Gaussian kernel function and $\mathbf{S}$ is similarity gram matrix. $\mathit{F}$ is the operation of setting the diagonal and non-topk elements of the gram matrix to 0, i.e. we select $k$ nearest neighbors in the whole dataset for each sample.

Ensuring the reliability of semantic information necessitates obtaining a high-quality feature representation, denoted as $\mathbf{z}$. While a straightforward approach involves extracting this representation from the model backbone online during training, our observations reveal that, particularly in the early stages of training, suboptimal classifiers can introduce noise during backbone updating. This noise can lead the model to learn in inappropriate directions, causing a degradation in feature quality throughout training.
To address this issue, we propose obtaining the feature representation from the initial pre-trained model backbone. This approach not only mitigates the problem but also reduces training time costs, as it necessitates computing the adjacency matrix only once at the start, as opposed to the ongoing construction of the K-NNG during training.

Then, we utilize the adjacency matrix $\mathbf{A}$ as a semantic regularization term. Intuitively, our goal is for samples with close proximity in the feature space also to exhibit low distances in the output and label spaces. The final formulation of our semantic-regularized progressive partial optimal transport (SP$^2$OT) is as follows:
\begin{align}\label{eq:sppot}
&\min_{\mathbf{Q} \in \Pi} \langle\mathbf{Q},-\log \mathbf P\rangle_F
 -  \lambda_1 \langle \mathbf{A}, \mathbf{Q}\mathbf{Q^\top} \rangle_F  + \lambda_2 KL(\mathbf{Q}^\top \bm1_N,\frac{\rho}{K}\bm 1_K) \\
 &\text{s.t.}\quad \Pi = \{\mathbf{Q} \in \mathbb{R}^{N\times K}_+ | \mathbf{Q} \bm1_K \leq\frac{1}{N}\bm1_N, \bm 1_N^\top\mathbf{Q} \bm 1_K=\rho\}, \label{eq:sppot_2}
\end{align}
where $\lambda_1$ is the regulation factor for semantic regularization. We let $\lambda_1$ decay with the training process in our implementation. The information from model prediction gradually becomes reliable, while the semantic information is from the initial pre-trained model backbone, which may limit the model's flexibility. Specifically, we reuse $\rho$, which increases with the training progress and will be illustrated in Sec.\ref{sec:rho}, to decay $\lambda_1$:
\begin{equation}\label{eq:lambda1}
    \lambda_1 = \lambda_1^0 \cdot (1-\rho),
\end{equation} 
where $\lambda_1^0$ is the initial value of $\lambda_1$.

\noindent\textbf{Relation to F-GWOT:}
To clarify the connection between our formulation and Fused Gromov-Wasserstein Optimal Transport (F-GWOT), we first revisit the notation introduced in Section~\ref{sec:fgwot}. Let $\mathcal{Z}$ denote the feature representation space and $\mathcal{Y}$ the class space. Here, $C(z_i, z_k)$ corresponds to the adjacency matrix $\mathbf{A}$, which encodes pairwise distances (or similarities) between samples in $\mathcal{Z}$, while $\hat{C}(y_j, y_l)$ represents the intrinsic relationships (e.g., semantic or structural) between classes in $\mathcal{Y}$.

Compared to the standard F-GWOT problem in Eq. (\ref{eq:gw_general_form}), our formulation differs in two key aspects: 1) We incorporate a total mass constraint, which progressively selects high-confidence samples during learning; 2) We discard cross-class interactions ($j \neq l$) in the term $\sum_{i,j,k,l} \mathcal{L}(C_{i,k}, \hat{C}{j,l}) \mathbf{Q}{i,j} \mathbf{Q}_{k,l}$, as these interactions are difficult to estimate reliably in deep clustering scenarios. This simplification ignores the intrinsic structure of $\mathcal{Y}$.

\section{Training}
To train our model, we initially generate pseudo-labels using our novel SP$^2$OT algorithm and subsequently utilize these pseudo-labels to supervise the model learning process. Moreover, inspired by curriculum learning, we progressively select hard samples for model learning by gradually increasing the parameter $\rho$. This section first introduces the increasing strategy of $\rho$ in Section \ref{sec:rho}.
We then detail the optimization algorithm for our novel SP$^2$OT problem in Section \ref{sec:SPPOT}, based on projected mirror descent. The algorithm proceeds iteratively through two alternating steps: (1) computation of the SP$^2$OT gradient, followed by (2) gradient descent with projection, where we formulate the projection step as a progressive partial optimal transport (P$^2$OT) problem.

\subsection{Training strategy of $\rho$}\label{sec:rho}
Selecting high-confidence samples for model learning is crucial in enhancing the quality of learned representations. To achieve that, we adopt a curriculum learning paradigm. Specifically, we select high-confidence samples by total mass constraint $\rho$ and gradually increase the value of $\rho$. In Eq.(\ref{eq:sppot}), instead of starting with 0 and incrementally increasing to 1, we introduce an initial value $\rho_0$ to mitigate potential issues associated with the model learning from very limited samples in the initial stages. We utilize the sigmoid ramp-up function, a technique commonly employed in semi-supervised learning~\cite{laine2016temporal,tarvainen2017mean}. Therefore, the increasing strategy of $\rho$ is expressed as follows:
\begin{equation} \label{eq:rho}
\rho = \rho_0 + (1 - \rho_0) \cdot e^{-5(1 - t/T)^2},
\end{equation}
where $T$ represents the total number of iterations, and $t$ represents the current iteration. We analyze and compare alternate design choices (e.g. linear ramp-up function) in our ablation study. Furthermore, we believe that a more sensible approach to setting $\rho$ involves adaptively increasing it based on the model's learning progress rather than relying on a fixed parameterization tied to the number of iterations. We leave the advanced design of $\rho$ for future work.

\subsection{Projected Mirror Descent for SP$^2$OT}\label{sec:SPPOT}

With the introduction of semantic regularization, Eq. (\ref{eq:sppot}) no longer admits a solution via scaling algorithms. To overcome this challenge, inspired by~\cite{peyre2016gromov}, we augment the objective with an entropy regularization term \(-\epsilon \mathcal{H}(\mathbf{Q})\) and adopt a projected mirror descent algorithm, which alternatively derive the gradient and gradient descent with projection until converge. The projection can be reformulated as an entropy-regularized progressive partial optimal transport (P\(^2\)OT) problem~\cite{peyre2016gromov,benamou2015iterative}.  

However, due to the total mass constraint, solving P²OT using generalized scaling algorithms (GSA) remains inefficient. To accelerate optimization, we propose a strategy that reformulates P\(^2\)OT into a form similar to Eq. (\ref{eq:general_form}), enabling faster convergence—particularly when \(\rho\) approaches 1—compared to GSA methods.


Specifically, the iteration of the projected mirror descent algorithm is given by:
\begin{align}
\mathbf{Q} \leftarrow \text{Proj}_{\text{KL}}^{\Pi,\text{KL}}\left( \mathbf{Q} \odot \exp\left( -\tau (\nabla f(\mathbf{Q}) - \varepsilon \nabla \mathcal{H}(\mathbf{Q})) \right) \right),
\end{align}
where $f(\mathbf{Q}) = \langle\mathbf{Q},-\log \mathbf P\rangle_F - \lambda_1 \langle \mathbf{A}, \mathbf{Q}\mathbf{Q^\top} \rangle_F$, $\text{Proj}_{\text{KL}}$ denotes the KL projector, and the superscript ($\Pi,\text{KL}$) of $\text{Proj}_{\text{KL}}$ indicates the constraints on $\mathbf{Q}$ specified in Eq.(\ref{eq:sppot}).

As demonstrated in~\cite{peyre2016gromov,benamou2015iterative}, the above projection is equivalent to the following entropy-regularized progressive partial optimal transport (P$^2$OT) problem:
\begin{align}\label{eq:sppot_mm}
\min_{\mathbf{Q} \in \Pi} \langle \mathbf{Q}, \mathbf{C}\rangle_F 
+ \lambda_2 \text{KL}(\mathbf{Q}^\top &\bm{1}_N, \tfrac{\rho}{K}\bm{1}_K) - \epsilon \mathcal{H}(\mathbf{Q}) \\
\text{s.t.}\quad \Pi = \{\mathbf{Q} \in \mathbb{R}^{N\times K}_+ \mid \mathbf{Q} &\bm{1}_K \leq \tfrac{1}{N}\bm{1}_N, \bm{1}_N^\top\mathbf{Q} \bm{1}_K = \rho\}
\end{align}
where $\mathbf{C} = \nabla f(\mathbf{Q}) = -\log \mathbf{P} - \lambda_1 (\mathbf{A}+\mathbf{A}^\top) \mathbf{Q}$. This formulation efficiently combines gradient descent with simultaneous constraint enforcement. The complete algorithm is presented in Algorithm~\ref{alg:sppot}.

We adopt the technique proposed by~\cite{zhang2024p2ot}, achieving faster convergence compared to the generalized scaling algorithm, particularly when $\rho$ is close to 1. We also note that since our SP$^2$OT is a special case of unbalanced Fused Gromov-Wasserstein Optimal Transport, our algorithm can be extended to solve the more general unbalanced Fused Gromov-Wasserstein Optimal Transport problem. Our primary contribution lies in extending~\cite{zhang2024p2ot} to solve this more complex transport problem. 

In the following section, we present our novel method for solving the P$^2$OT problem in detail.

\begin{algorithm}[t!]
    \caption{Semantic regularized pseudo label generation (SP$^2$OT)}\label{alg:sppot}
    \SetAlgoLined
    \footnotesize
    \DontPrintSemicolon
    \KwIn{Prediction probability matrix P ,adjacency matrix $ \mathbf{A}$, regulation weight $\lambda_1$, $\lambda_2$, $\rho$, $\epsilon$}    

    $\mathbf{C}_0 \leftarrow -\log \mathbf{P}$

    $\mathbf{Q} \leftarrow \mathbf{\frac{\rho}{N}}_{N\times K}$
    
    \While{$\bm Q$ not converge}{
    \Comment{Derive the gradient and treat it as cost}
    
    $\mathbf{C} \leftarrow \mathbf{C}_0 - \lambda_1 (\mathbf{A}+\mathbf{A^\top}) \mathbf{Q}$

    \Comment{Perform projection to update $\mathbf{Q}$}
    $\mathbf{Q} \leftarrow \text{P$^2$OT}(\mathbf{C}, \mathbf{Q}, \epsilon, \lambda_2, \rho)$
    }

    \KwRet $\mathbf{Q}$
    
\end{algorithm}

\subsection{Solver for P$^2$OT}\label{sec:solverppot}
Our approach involves introducing a virtual cluster onto the marginal~\cite{caffarelli2010free,chapel2020partial}. This virtual cluster serves the purpose of absorbing the $1 - \rho$ unselected mass, enabling us to transform the total mass constraint into the marginal constraint. Additionally, we replace the $KL$ penalty with a weighted $KL$ penalty to ensure strict adherence to the total mass constraint. As a result, we can reformulate P$^2$OT into a form akin to Eq.(\ref{eq:general_form}) and prove their solutions can be interconverted. Subsequently, we resolve this reformulated problem using an efficient scaling algorithm. As shown in Sec.\ref{sec:as}, compared to the generalized scaling solver proposed by ~\cite{chizat2018scaling}, our solver is two times faster.

Specifically, we denote the assignment of samples on the virtual cluster as $\bm\xi$. Then, we extend $\bm\xi$ to $\mathbf{Q}$, and denote the extended $\mathbf{Q}$ as $\hat{\mathbf{Q}}$ which satisfies the following constraints:
\begin{equation}\label{eq:extend}
    \hat{\mathbf{Q}} = [\mathbf{Q}, \bm\xi] \in \mathbb{R}^{N\times (K+1)}, \quad \bm\xi \in \mathbb{R}^{N\times 1}, \quad \hat{\mathbf{Q}}\bm 1_{K+1}=\bm 1_N .
\end{equation}
Due to $\bm 1_N^\top\mathbf{Q} \bm 1_K=\rho$, we known that,
\begin{equation}
    \bm 1_N^\top\hat{\mathbf{Q}} \bm 1_{K+1}=\bm 1_N^\top\mathbf{Q} \bm 1_K + \bm 1_N^\top\bm\xi=1 \Rightarrow \bm 1_N^\top\bm\xi = 1 - \rho .
\end{equation}
Therefore,
\begin{equation}\label{eq:row_constraint}
    \hat{\mathbf{Q}}^\top\bm 1_N = \left[\begin{array}{cc}
         \mathbf{Q}^\top\bm 1_N  \\
         \bm\xi^\top \bm 1_N 
    \end{array}\right] = \left[\begin{array}{cc}
         \mathbf{Q}^\top\bm 1_N  \\
         1 - \rho 
    \end{array}\right].
\end{equation}
The equation is due to $1_N^\top\bm\xi = \bm\xi^T \bm 1_N = 1-\rho$. We denote $\mathbf{C} = [f(\mathbf{Q}_t), \bm 0_N]$ and replace $\mathbf{Q}$ with $\hat{\mathbf{Q}}$, thus the Eq.(\ref{eq:curr_unbalanced_ot}) can be rewritten as follows:
\begin{align}\label{eq:re_curr_unbalanced_ot}
\min_{\hat{\mathbf{Q}} \in \Phi}\langle\hat{\mathbf{Q}},\mathbf{C}\rangle_F + \lambda KL(\hat{\mathbf{Q}}^\top &\bm1_N, \bm\beta) - \epsilon \mathcal{H}(\hat{\mathbf{Q}}) , \\
\text{s.t.}\ \Phi = \{\hat{\mathbf{Q}} \in \mathbb{R}^{N\times (K+1)}_+ | \hat{\mathbf{Q}}\bm1_{K+1}=&\frac{1}{N}\bm1_N\}, \bm\beta = \left[\begin{array}{cc}
         \frac{\rho}{K} \bm 1_K  \\
         1 - \rho 
    \end{array}\right].
\end{align}
However, due to the $KL$ penalty can not guarantee Eq.(\ref{eq:row_constraint}) is strictly satisfied, i.e. $\bm\xi^\top \bm 1_N = 1 - \rho$.
To solve this problem, we replace the standard $KL$ with weighted $KL$, which enables us to control the penalty strength for each class. The formula of weighted $KL$ is denoted as follows:
\begin{equation}
    \hat{KL}(\hat{\mathbf{Q}}^\top \bm1_N, \bm\beta, \bm\lambda) = \sum_{i=1}^{K+1} \bm\lambda_i [\hat{\mathbf{Q}}^\top\bm 1_N]_i \log \frac{[\hat{\mathbf{Q}}^\top\bm 1_N]_i}{\bm\beta_i} .
\end{equation}
Therefore, the Eq.(\ref{eq:curr_unbalanced_ot}) can be rewritten as follows:
\begin{align}\label{eq:re_curr_unbalanced_ot_2}
\min_{\hat{\mathbf{Q}} \in \Phi}\langle\hat{\mathbf{Q}},\mathbf{C}\rangle_F + \hat{KL}(\hat{\mathbf{Q}}^\top& \bm1_N, \bm\beta, \bm\lambda) - \epsilon \mathcal{H}(\hat{\mathbf{Q}}) \\
\text{s.t.}\quad \Phi = \{\hat{\mathbf{Q}} \in \mathbb{R}^{N\times (K+1)}_+ |& \hat{\mathbf{Q}}\bm1_{K+1}=\frac{1}{N}\bm1_N\}, \\ \quad \bm\beta = \left[\begin{array}{cc}
         \frac{\rho}{K} \bm 1_K  \\
         1 - \rho 
    \end{array}\right], \quad &\bm\lambda_{K+1} \rightarrow +\infty.
\end{align}
Intuitively, to assure Eq.(\ref{eq:row_constraint}) is strictly satisfied, we set $\bm\lambda_{K+1} \rightarrow +\infty$. In practice, as $\bm{\lambda}_{K+1} \rightarrow +\infty$ is equivalent to $\bm{f}$ in Algorithm \ref{alg:stable_srot} approaching $\bm{1}$, we directly set $\bm{f} = \bm{1}$ to avoid potential numerical instabilities. This places a substantial penalty on the virtual cluster, compelling the algorithm to assign a size of $1-\rho$ to the virtual cluster.


\begin{prop}\label{prop:equivalent}
If $\mathbf C = [f(\mathbf{Q}_t), \bm 0_N]$, and $\bm\lambda_{:K}=\lambda, \bm\lambda_{K+1} \rightarrow +\infty$, the optimal transport plan $\hat{\mathbf{Q}}^\star$ of Eq.(\ref{eq:re_curr_unbalanced_ot_2}) can be expressed as:
\begin{equation}
    \hat{\mathbf{Q}}^\star = [\mathbf{Q}^\star, \bm\xi^\star],
\end{equation}
where $\mathbf{Q}^\star$ is optimal transport plan of Eq.(\ref{eq:sppot_mm}), and $\bm\xi^\star$ is the last column of $\hat{\mathbf{Q}}^\star$.
\end{prop}

The proof is in the Appendix 3. Consequently, we focus on solving Eq.(\ref{eq:re_curr_unbalanced_ot_2}) to obtain the optimal $\hat{\mathbf{Q}}^\star$.

\begin{prop}\label{prop:solver}
Adding a entropy regularization $-\epsilon\mathcal{H}(\hat{\mathbf{Q}})$ to Eq.(\ref{eq:re_curr_unbalanced_ot_2}), we can solve it by efficient scaling algorithm. We denote $\mathbf M = \exp (-\mathbf{C}/\epsilon), \bm f= \frac{\bm\lambda}{\bm\lambda + \epsilon},\bm\alpha=\frac{1}{N}\bm1_N$. The optimal $\hat{\mathbf{Q}}^\star$ is denoted as follows:
\begin{equation}
    \hat{\mathbf{Q}}^\star = {\emph{diag}}(\mathbf{a}) \mathbf{M} \emph{diag}(\mathbf{b}),
\end{equation}
where $\mathbf{a,b}$ are two scaling coefficient vectors and can be derived by the following recursion formula:
\begin{equation}
    \mathbf{a} \leftarrow \frac{\bm{\alpha}}{\mathbf{M} \mathbf{b}}, \quad \mathbf{b} \leftarrow (\frac{\bm{\beta}}{\mathbf{M}^\top \mathbf{a}})^{\circ\bm f},
\end{equation}
where $\circ$ denotes Hadamard power, i.e., element-wise power. The recursion will stop until $\bm b$ converges.
\end{prop}
The proof is in the Appendix 4. Moreover, we employ the Log-domain stabilization trick in \cite{chizat2018scaling}, which absorbs extreme values in $\mathbf{a,b}$ and keeps them close to 1 to alleviate the numerically imprecise problem and potentially accelerate the convergence. The pseudo-code is shown in Algorithm \ref{alg:stable_srot}. The efficiency analysis is in the Sec.\ref{sec:as}.

In practical scenarios, solving P$^2$OT for the entire dataset is often impractical. Therefore, we adopt a more feasible approach by implementing mini-batch optimal transport. However, mini-batch OT may face challenges related to insufficient statistical representation due to the limitations imposed by the smaller subset of data. To address this limitation, we introduce a memory buffer that stores a substantial number of sample predictions (e.g., 5120) to ensure the existence of minority clusters. Specifically, before inputting data into P$^2$OT in each iteration, we concatenate predictions from the memory buffer with the current batch predictions to enhance stability while preserving efficiency.

In summary, the details of our training pipeline are outlined in Algorithm \ref{alg:overall}.


\SetKwComment{Comment}{//}{}









\begin{algorithm}[t!]
    \caption{Scaling Algorithm for P$^2$OT with stabilization}\label{alg:stable_srot}
    \SetAlgoLined
    \footnotesize
    \DontPrintSemicolon
    \KwIn{Cost matrix $-\log \mathbf P$, $\epsilon$, $\lambda$, $\rho$, $N,K$, a large value $\iota$}    
    $\mathbf{C} \leftarrow [-\log \mathbf P, \bm 0_N], \quad\bm \lambda \leftarrow [\lambda, ..., \lambda, \iota]^\top$
    
    $\bm \beta  \leftarrow [\frac{\rho}{K} \bm 1_K^\top, 1-\rho]^\top, \quad \bm{\alpha}  \leftarrow  \frac{1}{N}\mathbf{1}_N $
    
    $\mathbf{b} \leftarrow \mathbf{1}_{K+1}, \quad \mathbf{M} \leftarrow \exp(-\mathbf{C}/\epsilon), \quad \bm f \leftarrow \frac{\bm \lambda}{\bm\lambda + \epsilon}$\\
    $\mathbf{u} \leftarrow \mathbf{0}_N, \quad \mathbf{v} \leftarrow \mathbf{0}_{K+1}, \quad \mathbf{w} \leftarrow \exp(\frac{\mathbf{v}(\bm f-1)}{\epsilon} )$
    
    \While{$\bm b$ not converge}{
    
    $\mathbf{a} \leftarrow \frac{\bm{\alpha}}{\mathbf{M} \mathbf{b}}$
    
    $\mathbf{b} \leftarrow \mathbf{w}(\frac{\bm{\beta}}{\mathbf{M}^\top \mathbf{a}})^{\circ\bm f}  $
    
    \If {max(max($\mathbf{a}$),max($\mathbf{b}$)) is too large} {
        $(\mathbf{u},\mathbf{v})\leftarrow  (\mathbf{u}+\epsilon log(\mathbf{a}), \mathbf{v}+\epsilon log(\mathbf{b}))$ 
        
        $\mathbf{w} \leftarrow \mathbf{w}\mathbf{b}^{\bm f -1}$
        
        $\mathbf{M} \leftarrow \exp((\mathbf{u}-\mathbf{C}+\mathbf{v})/\epsilon)$
    
        $(\mathbf{b}, \mathbf{a}) \leftarrow (\mathbf{1}_{K+1}, \mathbf{1}_N) $
    }
    
    }
    
    $\mathbf{Q} \leftarrow \text{diag}(\mathbf{a}) \mathbf{M} \text{diag}(\mathbf{b}) $
    
    \KwRet $\mathbf{Q}[:, :K]$
    
\end{algorithm}

\begin{algorithm}[t!]
    \caption{Overall training strategy}\label{alg:overall}
    \SetAlgoLined
    \footnotesize
    \DontPrintSemicolon
    \KwIn{Training dataset $\mathcal{D}$, initial feature extractor (backbone) $f_0$, model $f_\theta$, stochastic transformation $T$, memory buffer, max\_epoch, $\rho_0$, $\lambda_1^0$, $\lambda_2$, $\epsilon$, $\eta$}

    $\mathbf{Z} \leftarrow f_0(\mathcal{D})$

    $\mathbf{A}$ $\leftarrow$ construct k-NNG from $\mathbf{Z}$ by Eq.(\ref{eq:adjacency1}, \ref{eq:adjacency2})
    
    \For{epoch = 1, ..., max\_epoch}{
        \For{iter = 1, ..., max\_iteration}{
            $\rho \leftarrow \rho_0 + (1 - \rho_0) \cdot e^{-5(1 - iter/max\_iteration)^2}$

            $\lambda_1 \leftarrow \lambda_1^0 \cdot (1-\rho)$

            $X, I = Sample(\mathcal{D})$

            $V_1, V_2 = T(X), T(X)$
            
            $\mathbf P_1, \mathbf P_2 = f_{\theta}(V_1), f_{\theta}(V_2)$

            \Comment{$I^\prime$ is the index of data.}
            
            $\mathbf{M}_1, \mathbf{M}_2, I^\prime = \text{MemoryBuffer}(\mathbf P_1, \mathbf P_2, I)$

            $\mathbf A^\prime = \mathbf{A}[I^\prime, I^\prime]$

            $\mathbf Q_1 = \mathrm{SP^2OT( \mathbf{M}_1, \mathbf A^\prime, \lambda_1, \lambda_2, \rho, \epsilon)}$

            $\mathbf Q_2 = \mathrm{SP^2OT(\mathbf{M}_2, \mathbf A^\prime, \lambda_1, \lambda_2, \rho, \epsilon)}$

            \Comment{Swap the prediction of two views.}
            
            $\mathcal{L} \leftarrow \langle \mathbf Q_2, -\log \mathbf P_1 \rangle_F + \langle \mathbf Q_1, -\log \mathbf P_2 \rangle_F$

            $\theta \leftarrow \theta - \eta \nabla_\theta \mathcal{L}$\\
        }
     }
     \KwRet $\theta$

\end{algorithm}

%% file: sec/exp.tex
\section{Experiments}
\begin{table*}[!t]
\caption{Comparison with SOTA methods on different imbalanced training sets. The best results are shown in boldface, and the next best results are indicated with an underscore.}
\LARGE

\label{tab:main}
\resizebox{1\textwidth}{!}{
\begin{tabular}{c|cccccccccccc}
\toprule
\multirow{2}{*}{Method} & \multicolumn{3}{c}{CIFAR100}                      & \multicolumn{3}{c}{ImgNet-R}                      & \multicolumn{3}{c}{iNature100}                    & \multicolumn{3}{c}{iNature500}                                       \\
        &ACC                          &NMI                          &F1                           &ACC                          &NMI                          &F1                           &ACC                          &NMI                          &F1                           &ACC                          &NMI                          &F1                           \\ \midrule
DINO & 36.6 & 68.9 & 31.0 & 20.5 & 39.6 & 22.2 & 40.1 & 67.8 & 34.2 & 29.8 & 67.0 & 24.0 \\
BCL & 35.7 & 66.0 & 29.9 & 20.7 & 40.0 & 22.4 & 41.9 & 67.2 &35.4 & 28.1 & 64.7 & 22.4 \\
IIC     &27.3$_{\pm 3.1}$             &65.0$_{\pm 1.8}$             &23.0$_{\pm 2.6}$             &18.7$_{\pm 1.5}$             &39.6$_{\pm 1.1}$             &15.9$_{\pm 1.5}$             &28.5$_{\pm 1.6}$             &63.9$_{\pm 1.0}$             &22.2$_{\pm 1.2}$             &13.1$_{\pm 0.3}$             &58.4$_{\pm 0.3}$             &7.1$_{\pm 0.2}$              \\
PICA    &29.8$_{\pm 0.6}$             &59.9$_{\pm 0.2}$             &24.0$_{\pm 0.2}$             &12.6$_{\pm 0.3}$             &34.0$_{\pm 0.0}$             &12.1$_{\pm 0.2}$             &34.8$_{\pm 2.4}$             &54.8$_{\pm 0.6}$             &23.8$_{\pm 1.2}$             &16.3$_{\pm 0.3}$             &55.9$_{\pm 0.1}$             &11.3$_{\pm 0.1}$             \\
SCAN    &37.2$_{\pm 0.9}$             &\underline{69.4}$_{\pm 0.4}$ &31.4$_{\pm 0.7}$             &21.8$_{\pm 0.7}$             &42.6$_{\pm 0.3}$             &21.7$_{\pm 0.8}$             &38.7$_{\pm 0.6}$             &66.3$_{\pm 0.5}$             &28.4$_{\pm 0.6}$             &29.0$_{\pm 0.3}$             &66.7$_{\pm 0.2}$             &21.6$_{\pm 0.2}$             \\
SCAN*   &30.2$_{\pm 0.9}$             &68.5$_{\pm 2.3}$             &25.4$_{\pm 1.1}$             &23.6$_{\pm 0.2}$             &44.1$_{\pm 0.2}$             &22.8$_{\pm 0.1}$             &39.5$_{\pm 0.4}$             &\textbf{68.5}$_{\pm 0.2}$    &30.7$_{\pm 0.1}$             &19.0$_{\pm 0.7}$             &65.9$_{\pm 0.5}$             &12.5$_{\pm 0.5}$             \\
CC      &29.0$_{\pm 0.6}$             &60.7$_{\pm 0.6}$             &24.6$_{\pm 0.5}$             &12.1$_{\pm 0.6}$             &30.5$_{\pm 0.1}$             &11.2$_{\pm 0.9}$             &28.2$_{\pm 2.4}$             &56.1$_{\pm 1.2}$             &20.6$_{\pm 2.1}$             &16.5$_{\pm 0.7}$             &55.5$_{\pm 0.0}$             &12.3$_{\pm 0.4}$             \\
DivClust&31.8$_{\pm 0.3}$             &64.0$_{\pm 0.4}$             &26.1$_{\pm 0.8}$             &14.8$_{\pm 0.2}$             &33.9$_{\pm 0.4}$             &13.8$_{\pm 0.2}$             &33.7$_{\pm 0.2}$             &59.3$_{\pm 0.5}$             &23.3$_{\pm 0.7}$             &17.2$_{\pm 0.5}$             &56.4$_{\pm 0.3}$             &12.5$_{\pm 0.4}$             \\
P$^2$OT &\underline{38.2}$_{\pm 0.8}$ &\textbf{69.6}$_{\pm 0.3}$    &\underline{32.0}$_{\pm 0.9}$ &\underline{25.9}$_{\pm 0.9}$ &\textbf{45.7}$_{\pm 0.5}$    &\underline{27.3}$_{\pm 1.4}$ &\underline{44.2}$_{\pm 1.2}$ &67.0$_{\pm 0.6}$             &\underline{36.9}$_{\pm 2.0}$ &\underline{32.2}$_{\pm 2.0}$ &\textbf{67.2}$_{\pm 0.3}$    &\underline{25.2}$_{\pm 1.7}$ \\
SP$^2$OT&\textbf{39.1}$_{\pm 1.5}$    &67.6$_{\pm 0.8}$             &\textbf{32.8}$_{\pm 1.3}$    &\textbf{27.1}$_{\pm 0.6}$    &\underline{44.9}$_{\pm 0.7}$ &\textbf{29.1}$_{\pm 0.6}$    &\textbf{49.0}$_{\pm 1.3}$    &\underline{68.1}$_{\pm 0.4}$ &\textbf{41.8}$_{\pm 0.8}$    &\textbf{34.1}$_{\pm 0.8}$    &\underline{67.2}$_{\pm 0.9}$ &\textbf{27.1}$_{\pm 1.1}$    \\
\bottomrule
\end{tabular}}

\end{table*}

\begin{figure*}[!t]
    \centering
    \includegraphics[scale=0.43]{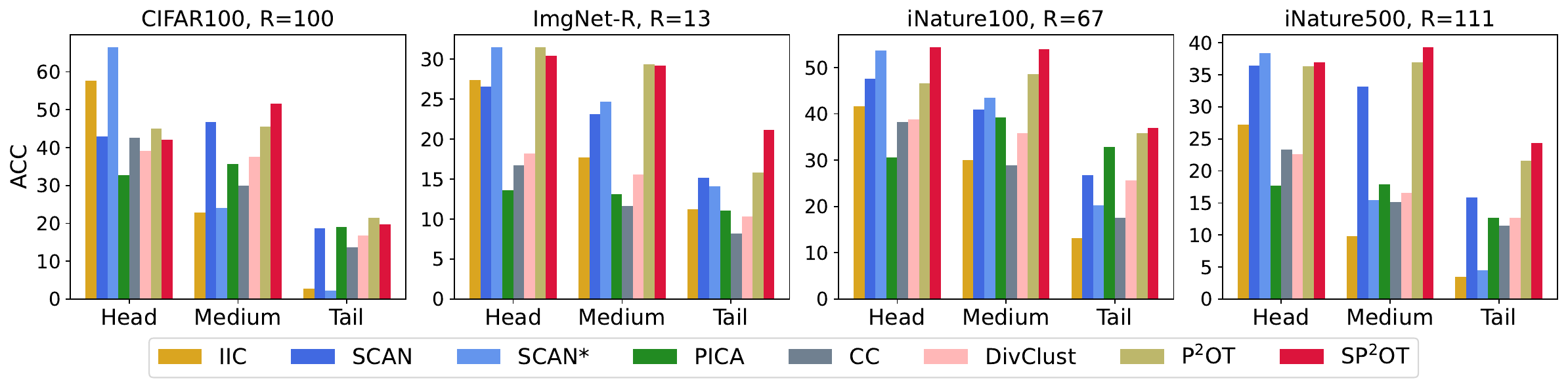}
    
    \caption{Head, Medium, and Tail comparison on several datasets.}
    \label{fig:hmt}
\end{figure*}

\begin{figure*}[!t]
    \centering
    \includegraphics[scale=0.45]{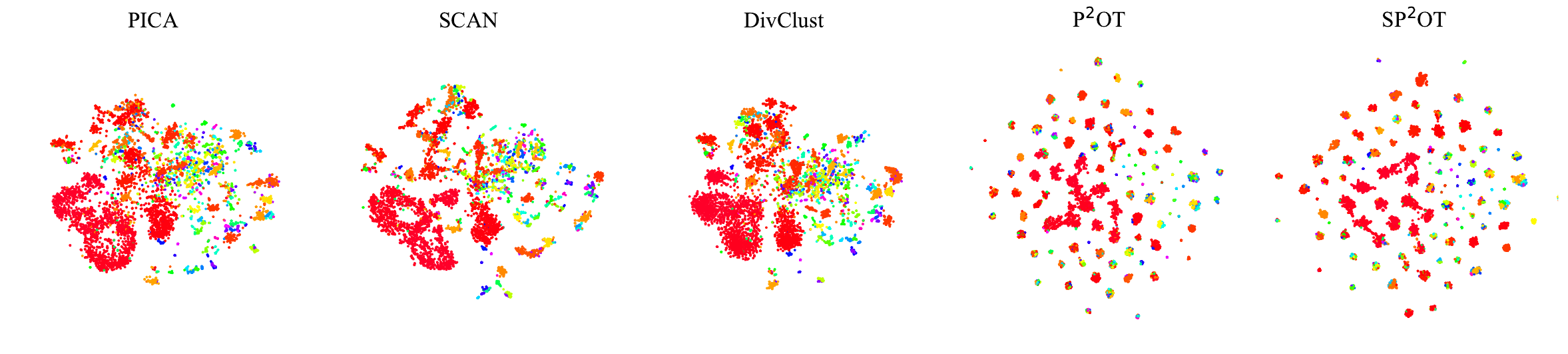}
    \vspace{-0.5em}
    \caption{The T-SNE analysis on iNature100 training set. The primary distinction between them lies in the distribution of the head classes, represented in red (best viewed when zoomed in on screen).}
    \label{fig:tsne}
    
\end{figure*}

\subsection{Experimental Setup}

\subsubsection{Datasets} 
To evaluate our method, we have established a realistic and challenging benchmark, including CIFAR100~\cite{krizhevsky2009learning}, ImageNet-R (abbreviated as ImgNet-R) ~\cite{hendrycks2021many} and iNaturalist2018~\cite{van2018inaturalist} datasets. To quantify the level of class imbalance, we introduce the imbalance ratio denoted as $R$, calculated as the ratio of $N_{max}$ to $N_{min}$, where $N_{max}$ represents the largest number of images in a class, and $N_{min}$ represents the smallest. For CIFAR100, as in \cite{cao2019learning}, we artificially construct a long-tailed CIFAR100 dataset with an imbalance ratio of 100. For ImgNet-R, which has renditions of 200 classes resulting in 30k images and is inherently imbalanced, we split 20 images per class as the test set, leaving the remaining data as the training set ($R=13$). Note that the data distribution of ImgNet-R is different from the ImageNet, which is commonly used for training unsupervised pre-trained models, posing a great challenge to its clustering. Consequently, ImgNet-R serves as a valuable resource for assessing the robustness of various methods. Furthermore, we incorporate the iNaturalist2018 dataset, a natural long-tailed dataset frequently used in supervised long-tailed learning~\cite{cao2019learning,zhang2021distribution}. This dataset encompasses 8,142 classes, posing significant challenges for clustering. To mitigate this complexity, we extract subsets of 100, 500, and 1000 classes, creating the iNature100 ($R=67$), iNature500 ($R=111$), and iNature1000 ($R=111$) datasets, respectively. iNature100 is the subset of iNature500, and iNature500 is the subset of iNature1000. The distribution of datasets is in the Appendix 69. We perform evaluations on both the imbalanced training set and the corresponding balanced test set. Note that we do not conduct experiments on ImageNet datasets because the unsupervised pretrained models have trained on the whole balanced ImageNet. 

\subsubsection{Evaluation Metric} 
We evaluate our method using the clustering accuracy (ACC) metric averaged over classes, normalized mutual information (NMI), and F1-score. We also provide the adjusted Rand index (ARI) metrics, which is an instance-wise evaluation metric and is not a suitable metric for imbalanced datasets, in the Appendix 8. To provide a more detailed analysis, we rank the classes by size in descending order and divide the dataset into Head, Medium, and Tail categories, maintaining a ratio of 3:4:3 across all datasets. Then, we evaluate performance on the Head, Medium, and Tail, respectively.


\begin{table*}[t]
\caption{Comparison with SOTA methods on different balanced test sets to evaluate generalization ability. The best results are shown in boldface, and the next best results are indicated with an underscore.}
\LARGE

\label{tab:main-testset}
\resizebox{1\textwidth}{!}{
\begin{tabular}{c|cccccccccccc}
\toprule
\multirow{2}{*}{Method} & \multicolumn{3}{c}{CIFAR100}                      & \multicolumn{3}{c}{ImgNet-R}                      & \multicolumn{3}{c}{iNature100}                    & \multicolumn{3}{c}{iNature500}                                       \\
&ACC                          &NMI                          &F1                           &ACC                          &NMI                          &F1                           &ACC                          &NMI                          &F1                           &ACC                          &NMI                          &F1                           \\ \midrule
IIC     &29.4$_{\pm 3.0}$             &56.0$_{\pm 1.6}$             &21.0$_{\pm 3.6}$             &21.4$_{\pm 1.7}$             &53.9$_{\pm 1.0}$             &17.7$_{\pm 1.9}$             &35.1$_{\pm 2.7}$             &75.4$_{\pm 1.4}$             &25.7$_{\pm 2.8}$             &19.8$_{\pm 0.7}$             &73.9$_{\pm 0.4}$             &10.6$_{\pm 0.4}$             \\
PICA    &30.3$_{\pm 0.4}$             &54.5$_{\pm 0.5}$             &28.3$_{\pm 0.6}$             &16.6$_{\pm 0.1}$             &51.8$_{\pm 0.1}$             &16.6$_{\pm 0.2}$             &43.6$_{\pm 3.3}$             &78.2$_{\pm 1.5}$             &38.7$_{\pm 3.5}$             &31.4$_{\pm 0.3}$             &79.6$_{\pm 0.2}$             &28.3$_{\pm 0.2}$             \\
SCAN    &37.5$_{\pm 0.3}$             &\underline{63.0}$_{\pm 0.1}$ &34.9$_{\pm 0.4}$             &23.8$_{\pm 0.8}$             &57.0$_{\pm 0.4}$             &23.8$_{\pm 0.9}$             &44.0$_{\pm 0.9}$             &80.8$_{\pm 0.5}$             &36.5$_{\pm 1.1}$             &39.0$_{\pm 0.1}$             &83.4$_{\pm 0.1}$             &33.3$_{\pm 0.3}$             \\
SCAN*   &30.5$_{\pm 0.6}$             &57.6$_{\pm 2.0}$             &19.8$_{\pm 0.9}$             &25.3$_{\pm 0.1}$             &57.1$_{\pm 0.3}$             &23.5$_{\pm 0.2}$             &44.8$_{\pm 0.7}$             &81.9$_{\pm 0.6}$             &36.8$_{\pm 1.3}$             &24.3$_{\pm 0.6}$             &77.0$_{\pm 0.6}$             &14.4$_{\pm 0.4}$             \\
CC      &29.1$_{\pm 1.3}$             &54.2$_{\pm 0.6}$             &25.6$_{\pm 1.3}$             &14.7$_{\pm 0.2}$             &48.5$_{\pm 0.3}$             &14.1$_{\pm 0.4}$             &38.2$_{\pm 1.4}$             &76.3$_{\pm 0.3}$             &32.4$_{\pm 0.6}$             &29.7$_{\pm 1.1}$             &78.3$_{\pm 0.7}$             &26.5$_{\pm 1.4}$             \\
DivClust&31.8$_{\pm 0.6}$             &58.1$_{\pm 0.3}$             &28.9$_{\pm 0.6}$             &16.8$_{\pm 0.3}$             &50.4$_{\pm 0.3}$             &16.4$_{\pm 0.3}$             &42.4$_{\pm 1.2}$             &78.7$_{\pm 0.6}$             &36.1$_{\pm 1.4}$             &31.6$_{\pm 0.5}$             &79.2$_{\pm 0.5}$             &27.8$_{\pm 0.6}$             \\
P$^2$OT &\underline{38.9}$_{\pm 1.1}$ &\textbf{63.1}$_{\pm 0.8}$    &\underline{35.9}$_{\pm 0.8}$ &\underline{27.5}$_{\pm 1.2}$ &\underline{58.0}$_{\pm 0.7}$ &\underline{28.1}$_{\pm 1.3}$ &\underline{50.0}$_{\pm 2.1}$ &\underline{83.0}$_{\pm 0.9}$ &\underline{44.2}$_{\pm 2.1}$ &\underline{42.2}$_{\pm 1.1}$ &\underline{84.3}$_{\pm 0.1}$ &\underline{37.2}$_{\pm 1.4}$ \\
SP$^2$OT&\textbf{39.0}$_{\pm 1.7}$    &61.8$_{\pm 1.3}$             &\textbf{36.6}$_{\pm 1.4}$    &\textbf{28.3}$_{\pm 0.4}$    &\textbf{58.1}$_{\pm 0.2}$    &\textbf{29.9}$_{\pm 0.4}$    &\textbf{53.8}$_{\pm 0.8}$    &\textbf{83.8}$_{\pm 0.3}$    &\textbf{48.1}$_{\pm 1.0}$    &\textbf{43.6}$_{\pm 0.3}$    &\textbf{84.7}$_{\pm 0.2}$    &\textbf{39.0}$_{\pm 0.2}$    \\
    \bottomrule
\end{tabular}}
\vspace{-0.5em}
\end{table*}

\subsubsection{Implementation Details} 
Building upon the advancements in transformer~\cite{dosovitskiy2020image} and unsupervised pre-trained models~\cite{he2022masked}, we conduct experiments on the ViT-B16, which is pre-trained with DINO~\cite{caron2021emerging}. To provide a comprehensive evaluation, we re-implement a variety of methods from existing literature, encompassing both deep clustering and imbalanced self-supervised learning approaches. Specifically, in the deep clustering domain, we report results for typical methods such as IIC \cite{ji2019invariant}, PICA \cite{huang2020pica}, CC \cite{li2021contrastive}, SCAN \cite{vangansbeke2020scan}, the strong two-stage SCAN* \cite{vangansbeke2020scan} which builds upon SCAN for self-labeling learning, SPICE \cite{niu2022spice}, and the recently proposed DivClust \cite{divclust2023}. In the realm of imbalanced self-supervised learning, we implement BCL \cite{zhou2022contrastive}, specifically designed for representation learning with long-tailed data. After representation learning, we proceed to perform clustering in the representation space. Notably, we exclude the re-implementation of \cite{jiang2021self} as they prune the ResNet, making it challenging to transfer to ViT.
It is important to note that all of these methods are trained using the same backbone, data augmentation, and training configurations to ensure a fair comparison. Specifically, we train 50 epochs and adopt the Adam optimizer with the learning rate decay from 5e-4 to 5e-6 for all datasets. The batch size is 512. Further details can be found in the Appendix 7. For hyperparameters, we set $\lambda$ as 1, $\epsilon$ as 0.1, and initial $\rho$ as 0.1. The stop criterion of Alg.3 is when the change of $\mathbf{b}$ is less than 1e-6, or the iteration reaches 1000. We utilize the loss on training sets for model selection. For evaluation, we conduct experiments with each method three times and report the mean results.


\subsection{Main Results} 
\subsubsection{Evaluation on Imbalanced Training Set} \label{main_results}
In Tab.\ref{tab:main}, we provide a comprehensive comparison of our method with existing approaches on various imbalanced training sets. On the relatively small-scale CIFAR100 dataset, our SP$^2$OT outperforms the previous state-of-the-art by achieving an increase of 0.9 in ACC, and 0.8 in F1 score. But 2.0\% decrease on the NMI metric. 
On the ImgNet-R datasets, our SP$^2$OT exhibits notable improvements compared to our previous P$^2$OT, with significant increases of 1.2 in accuracy (ACC) and 1.8 in F1 score. This underscores its effectiveness and robustness, particularly in out-of-distribution scenarios. While our method achieves superior clustering performance, as evidenced by higher ACC or F1 scores, it also leads to a relatively uniform distribution, which increases the denominator of the NMI metric and results in a decrease in the NMI metric. A detailed explanation can be found in the Appendix 9.

When applied to the fine-grained iNature datasets, our approach consistently delivers substantial performance gains across each subset in terms of ACC and F1 scores. Specifically, on ACC, we achieve improvements of 4.8 on iNature100 and 1.9 on iNature500. On the F1 score, we obtain improvements of 4.9 and 1.9 on the two datasets, respectively. In terms of NMI, compared to P$^2$OT, we observe a 1.1 improvement on iNature100, but compared to SCAN*, there is still a decrease of 0.4. It's worth noting that our method is an efficient one-stage approach, in contrast to SCAN*, which is a two-stage pseudo-labeling-based method. Additionally, another pseudo-labeling-based method, SPICE, exhibits a degenerate solution in the imbalance scenario (see Appendix 7). These results indicate that naive pseudo-labeling methods encounter significant challenges, emphasizing our method's superiority in handling imbalance scenarios.



In addition, we provide a detailed analysis of the results for the Head, Medium, and Tail classes, offering a more comprehensive understanding of our method's performance across different class sizes. As depicted in Fig. \ref{fig:hmt}, our improvements are predominantly driven by the Medium and Tail classes, especially in challenging scenarios like ImgNet-R and iNature500, although our results show some reductions in performance for the Head classes. These results highlight the effectiveness of our SP$^2$OT algorithm in generating imbalance-aware pseudo-labels, making it particularly advantageous for the Medium and Tail classes. 

Furthermore, in Fig. \ref{fig:tsne}, we present a T-SNE \cite{van2008visualizing} comparison of features before the clustering head. The T-SNE results illustrate that our method learns more distinct clusters, particularly benefiting Medium and Tail classes. In particular, in comparison to P$^2$OT, the head cluster in SP$^2$OT (depicted in red) is more separable.

\subsubsection{Evaluation on Balanced Test Set} \label{sec:balanced}
To assess the generalization ability of our method in an inductive setting, we conducted experiments on the balanced test sets of the three datasets. We did not compare with DINO and BCL on the test sets because they are not designed for transductive settings. As shown in Table \ref{tab:main-testset}, on CIFAR100, our method slightly outperforms P$^2$OT in terms of ACC and F1 score, with a decrease (1.3) in NMI. This decrease in NMI may be attributed to the fact that SP$^2$OT results in a more uniform distribution than P$^2$OT on CIFAR100. Notably, SP$^2$OT outperforms the state-of-the-art P$^2$OT on ImgNet-R, iNature100, and iNature500 by a significant margin, demonstrating its robustness in out-of-distribution scenarios and fine-grained scenarios.

\subsubsection{Evaluation on Large-scale iNature1000}  \label{sec:inature1000}
To further evaluate the scalability of our method, we conducted experiments on the large-scale iNature1000 dataset, which poses a significant challenge. As shown in Table \ref{tab:inature1000}, our method achieves the best performance on both training and test sets across all metrics. Specifically, our method outperforms the previous state-of-the-art method, P$^2$OT, by 1.3 in ACC, 1.2 in NMI, and 1.7 in F1 score on the training set. On the test set, our method achieves improvements of 1.2, 0.4, and 1.0 in ACC, NMI, and F1 scores, respectively. These results underscore the scalability of our method in large-scale scenarios.

\begin{table}[t]
\caption{Comparison with SOTA methods on iNature1000. The best results are shown in boldface, and the next best results are indicated with an underscore.}
\LARGE
\label{tab:inature1000}
\resizebox{0.48\textwidth}{!}{
\begin{tabular}{c|ccc|ccc}
\toprule
\multirow{2}{*}{Method} & \multicolumn{3}{c|}{Train}     & \multicolumn{3}{c}{Test}                                  \\
&ACC                          &NMI                          &F1                                      &ACC                          &NMI                          &F1\\ \midrule
IIC     &7.8$_{\pm 0.5}$              &56.9$_{\pm 0.4}$             &4.2$_{\pm 0.2}$              &13.3$_{\pm 1.0}$             &71.2$_{\pm 0.6}$             &5.1$_{\pm 0.7}$             \\
PICA    &12.4$_{\pm 0.2}$             &55.0$_{\pm 0.1}$             &8.5$_{\pm 0.1}$              &28.3$_{\pm 0.2}$             &80.0$_{\pm 0.2}$             &26.0$_{\pm 0.1}$            \\
SCAN    &26.2$_{\pm 0.2}$             &\underline{68.0}$_{\pm 0.0}$ &19.6$_{\pm 0.1}$             &36.5$_{\pm 0.3}$             &84.2$_{\pm 0.1}$             &31.7$_{\pm 0.4}$            \\
SCAN*   &10.5$_{\pm 0.2}$             &63.9$_{\pm 0.5}$             &6.4$_{\pm 0.2}$              &14.5$_{\pm 0.2}$             &72.8$_{\pm 0.3}$             &5.8$_{\pm 0.2}$             \\
CC      &12.1$_{\pm 0.4}$             &55.2$_{\pm 0.4}$             &9.1$_{\pm 0.4}$              &26.6$_{\pm 0.5}$             &78.7$_{\pm 0.2}$             &24.0$_{\pm 0.8}$            \\
DivClust&13.0$_{\pm 0.2}$             &56.0$_{\pm 0.3}$             &9.5$_{\pm 0.2}$              &26.9$_{\pm 0.1}$             &78.8$_{\pm 0.3}$             &24.3$_{\pm 0.2}$            \\
P$^2$OT &\underline{26.7}$_{\pm 0.5}$ &67.4$_{\pm 0.4}$             &\underline{19.9}$_{\pm 0.4}$ &\underline{37.7}$_{\pm 0.2}$ &\underline{84.4}$_{\pm 0.2}$ &\underline{33.0}$_{\pm 0.2}$\\
SP$^2$OT&\textbf{28.0}$_{\pm 0.8}$    &\textbf{68.6}$_{\pm 0.1}$    &\textbf{21.6}$_{\pm 0.9}$    &\textbf{38.9}$_{\pm 1.0}$    &\textbf{84.8}$_{\pm 0.3}$    &\textbf{34.0}$_{\pm 1.0}$   \\
 \bottomrule
\end{tabular}}
\vspace{-0.5em}
\end{table}

\vspace{-0.5em}
\subsection{Ablation Study}\label{sec:as}

\subsubsection{Component Analysis}
\begin{table*}[t]
\centering
\caption{Analysis of different formulations. SLA is proposed by ~\cite{tai2021sinkhorn} in semi-supervised learning. OT~\cite{asano2020self}, POT, UOT, and P$^2$OT are variants of our SP$^2$OT. P$^2$OT removes the semantic relation regularization in SP$^2$OT. POT substitutes the $KL$ constraint in P$^2$OT with the equality constraint. UOT removes the progressive $\rho$ in P$^2$OT.}

\label{tab:formulation}
\resizebox{1\textwidth}{!}{
\begin{tabular}{c|cccc|cccc|cccc}
\toprule
\multirow{2}{*}{Formulation} & \multicolumn{4}{c|}{CIFAR100}  & \multicolumn{4}{c|}{ImgNet-R}  & \multicolumn{4}{c}{iNature500} \\ 
                  & Head  & Medium & Tail  & ACC   & Head  & Medium & Tail  & ACC   & Head  & Medium & Tail  & ACC   \\\midrule
SLA                & - & -  & - & 3.26 & - & -  & - & 1.00 & - & -   & - & 6.33 \\
OT                 & 35.0 &  30.2 & 18.7 & 28.2 & 24.4  & 19.3 & 10.7 & 18.2 & 27.1 & 27.5 & 16.4 & 24.1\\
POT                & 39.6 &  42.9 & 14.5 & 33.4 & 25.5  & 23.3 & 17.6 &  22.2 & 31.7 & 32.7 & 19.7 &28.5  \\
UOT                & 43.4 &  42.5 & 19.6 & 35.9 & 27.7  & 23.9 & 14.6 & 22.2  & 32.2 & 29.0 & 18.3 & 26.7  \\
P$^2$OT            & 45.1 &  45.5 & 21.6 & 38.2 & 31.5  & 29.3 & 16.1 & 25.9  & 36.3 & 37.0 & 21.6 & 32.2  \\ 
SP$^2$OT           & 42.0 &  51.6 & 19.8 & 39.1 & 30.4  & 29.2 & 21.2 & 27.1  & 36.9 & 39.3 & 24.4 & 34.1 \\
\bottomrule
\end{tabular}}
\vspace{-0.5em}
\end{table*}

%
%

%
%

\begin{figure*}[!t]
    \centering
    \includegraphics[scale=0.48]{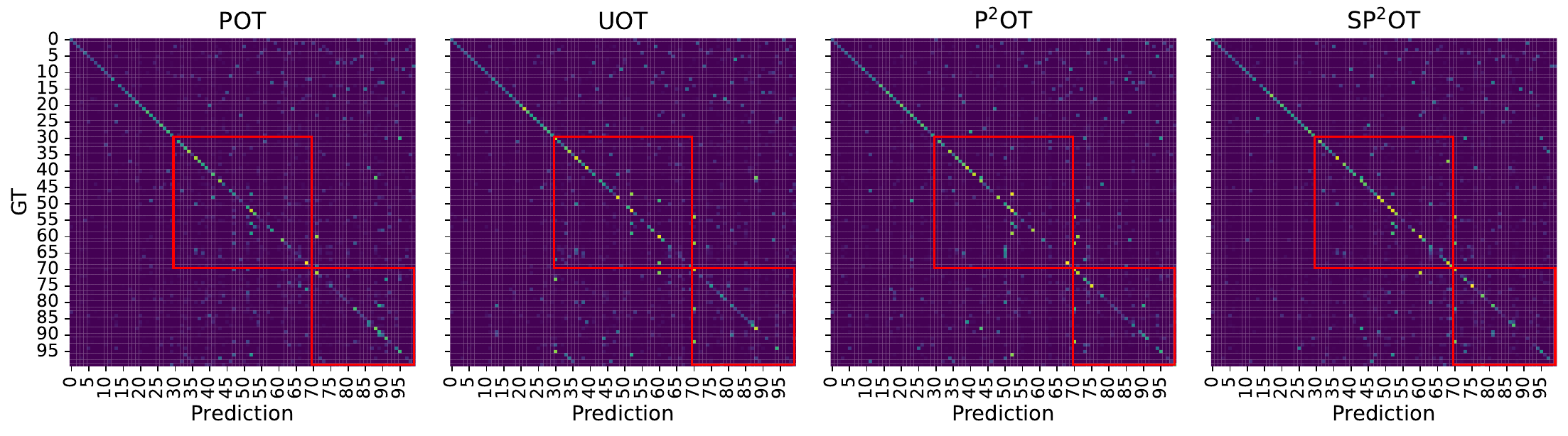}
    \caption{Confusion matrix on the balanced CIFAR100 test set. The two red rectangles represent the Medium and Tail classes.}
    \label{fig:confusion}
     
\end{figure*}


\begin{figure*}[!t]
    \centering
    \includegraphics[scale=0.5]{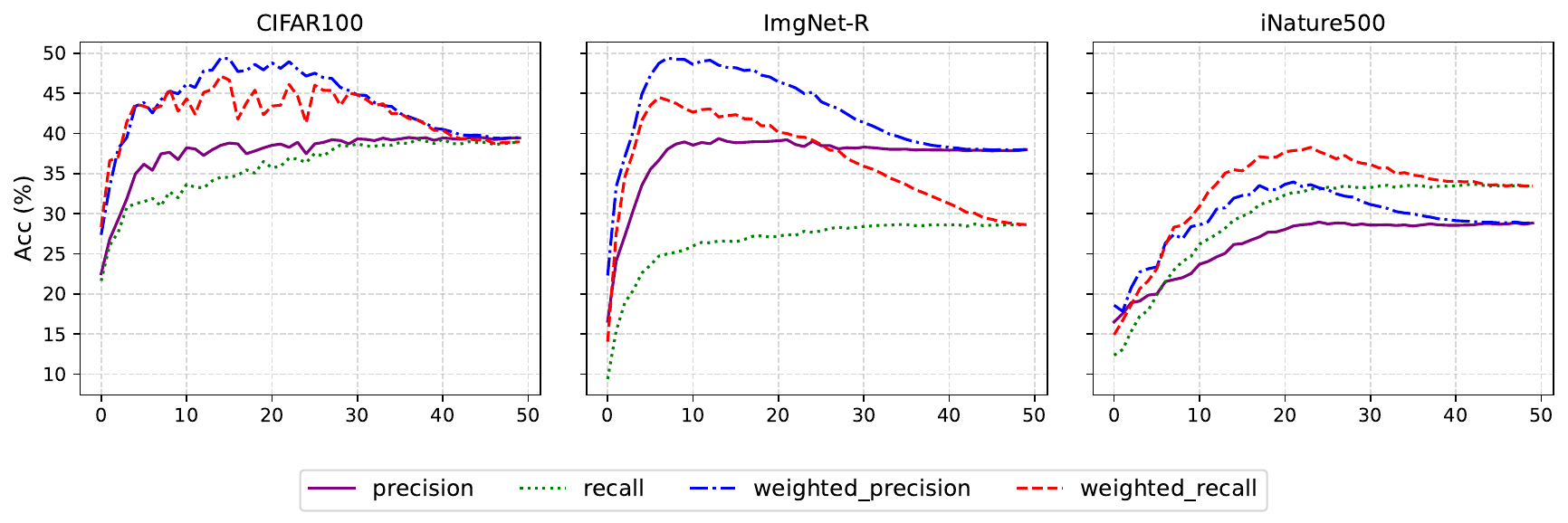}
    \vspace{-0.5em}
    \caption{Precision, Recall analysis on train dataset with different training epoch. The weighted precision and recall are derived by reweighting each sample by our SP$^2$OT algorithm.}
    \label{fig:pr}
\end{figure*}
To assess the $KL$ constraint, the progressive $\rho$, and the semantic regularization, we conduct ablation experiments to analyze their individual contributions. As shown in Table \ref{tab:formulation}, we compare our full SP$^2$OT method with three ablated versions: OT, POT, UOT, and P$^2$OT. OT ~\cite{asano2020self} imposes equality constraint on cluster size and without the progressive $\rho$ component. POT signifies the replacement of the $KL$ term in the P$^2$OT formulation with a typical uniform constraint. UOT denotes the removal of the progressive $\rho$ component from the P$^2$OT formulation. And P$^2$OT denotes SP$^2$OT to remove the semantic regularization. Furthermore, we also compare with
SLA~\cite{tai2021sinkhorn}, which relaxes the equality constraint of POT by utilizing an upper bound. However, due to this relaxation, samples are erroneously assigned to a single cluster in the early stages, rendering its failure for clustering imbalanced data. We detail their formulation and further analysis in Appendix 11. The results show POT and UOT both significantly surpass the OT, indicating that the $KL$ constraint and the progressive $\rho$ components yield satisfactory effects. Compared to POT, P$^2$OT achieves improvements of 5.8, 3.8, and 3.6 on CIFAR100, ImgNet-R, and iNature500, respectively. The improvement is mainly from Head and Tail classes, demonstrating that P$^2$OT with $KL$ constraint can generate imbalanced pseudo labels. Compared to UOT, P$^2$OT realizes gains of 2.3, 3.8, and 5.4 on CIFAR100, ImgNet-R, and iNature500, respectively. SP$^2$OT, which equips P$^2$OT with a novel semantic relation regularization, further improves P$^2$OT by a sizeable margin, and the improvements are mainly from Medium or Tail classes.

Additionally, as depicted in Fig.\ref{fig:confusion}, SP$^2$OT exhibits better results on medium classes, whereas P$^2$OT demonstrates better results on head classes.
\begin{figure}[!tp]
    \centering
    \includegraphics[scale=0.42]{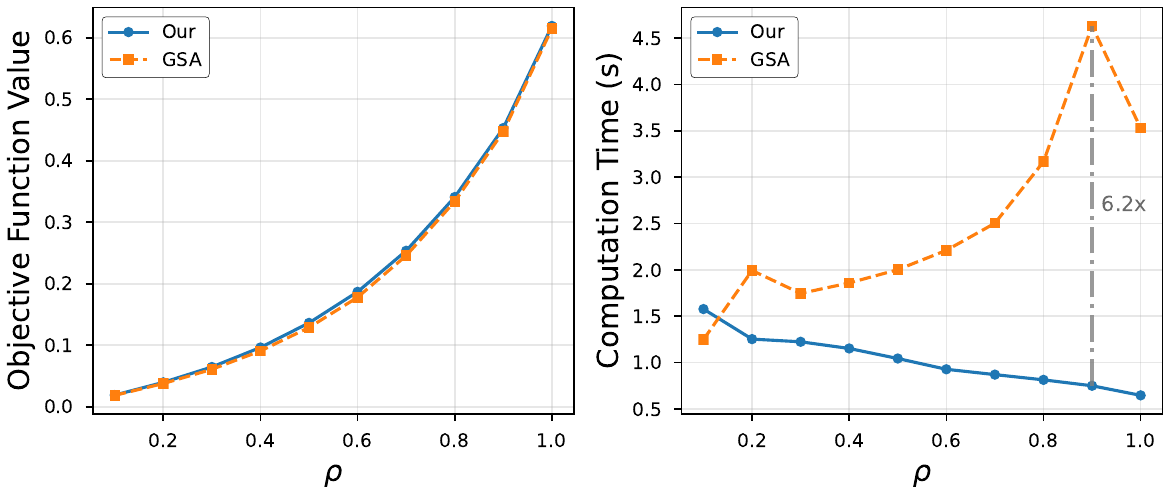}
    \caption{Comparison of our solver with Generalized Scaling Algorithm (GSA). The left figure is the objective function value and the right figure is the computation time of different methods.}
    \label{fig:time}
    
\end{figure}

\begin{table*}[!t]
\caption{Analysis of $\rho_0$ and different ramp up function. Fixed denotes $\rho$ as a constant value.}

\label{tab:rho}
\centering
\resizebox{0.70\textwidth}{!}{
\begin{tabular}{c|ccccc|ccc|cc}
\toprule
\multirow{2}{*}{$\rho_0$}   & \multicolumn{5}{c|}{Sigmoid} & \multicolumn{3}{c|}{Linear} & \multicolumn{2}{c}{Fixed} \\
& 0.00  & 0.05  & 0.1   & 0.15  & 0.2              & 0   & 0.1  & 0.2 & 0.1 & 0.2 \\ \midrule
CIFAR100   & 32.9 & 37.9 & 38.2 & 36.3 & 37.6 & 35.0 & 36.6 & 36.8 & 37.4 & 37.1 \\
ImgNet-R   & 26.5 & 27.2 & 25.9 & 26.0 & 26.6 & 25.8 & 28.5 & 23.0 & 25.5 & 24.7 \\
iNature500 & 32.6 & 31.7 & 32.2 & 32.8 & 32.9 & 31.9 & 32.0 & 30.5 & 29.9 & 30.3 \\ \bottomrule
\end{tabular}}
\vspace{-0.5em}
\end{table*}

\vspace{-0.3em}
\subsubsection{Pseudo Label Quality Analysis}
To provide a deeper understanding, we conduct an analysis of the pseudo-label quality generated by our P$^2$OT. We evaluate precision and recall metrics to assess the pseudo-label quality for the entire training set. Notably, our P$^2$OT algorithm conducts selection through reweighting, rather than hard selection. Consequently, we reweight each sample and present the weighted precision and recall results. As depicted in Fig.\ref{fig:pr}, our weighted precision and recall consistently outperform precision and recall across different epochs. In the early stages, weighted precision and recall exhibit more rapid improvements compared to precision and recall, demonstrating the effectiveness of our selection strategy. However, they eventually plateau around 10 epochs ($\rho \approx 0.15$), converging gradually with precision and recall. The decline observed in weighted precision and recall over time suggests that our current ramp-up function may not be optimal, and raising to 1 for $\rho$ may not be necessary. We believe that the ramp-up strategy for $\rho$ should be adaptive to the model's learning progress. In this paper, we have adopted a typical sigmoid ramp-up strategy, and leave more advanced designs for future work.

\vspace{-0.3em}
\subsubsection{Efficiency Analysis}
To demonstrate the efficiency of our solver, we perform a comparison between our solver and PMD with the Generalized Scaling Algorithm (GSA) proposed by \cite{chizat2018scaling} for SP$^2$OT. The pseudo-code of GSA is in the Appendix 12. This comparison is conducted on iNaure1000 using identical conditions (NVIDIA TITAN RTX, $\epsilon = 0.1, \lambda=1$), employing stabilization strategies for both.
To ensure a comprehensive analysis, we vary the parameter $\rho$ and compare both the time cost and the objective value. The experiments are conducted on the iNat500 dataset, which consists of 27K data points.  
The results, presented in Fig.~\ref{fig:time}, reveal several key insights:  
1) Under the same initialization, both methods are converged to nearly identical values.  
2) Our method achieves up to 6.2$\times$ speedup compared with GSA.
These findings demonstrate the efficiency of our solver, as proposed in Sec.~\ref{sec:solverppot}.



\vspace{-0.3em}
\subsubsection{Analysis of $\rho$}
The choice of initial $\rho_0$ and the specific ramp-up strategy are important hyperparameters. Therefore, in this section, we systematically investigate the impact of varying $\rho_0$ values and alternative ramp-up strategy on P$^2$OT, which removes the semantic regularization of SP$^2$OT and is more efficiency to train. The term "Linear" signifies that $\rho$ is increased to 1 from $\rho_0$ using a linear function, while "Fixed" indicates that $\rho$ remains constant as $\rho_0$. The results in Table \ref{tab:rho} provide several important insights: 1) Our method exhibits consistent performance across various $\rho_0$ values when using the sigmoid ramp-up function, highlighting its robustness on datasets like ImgNet-R and iNature500; 2) The linear ramp-up strategy, although slightly less effective than sigmoid, still demonstrates the importance of gradually increasing $\rho$ during the early training stage; 3) The fixed $\rho$ approach results in suboptimal performance, underscoring the necessity of having a dynamically increasing $\rho$. These findings suggest that a good starting value for $\rho_0$ is around 0.1, and it should be progressively increased during training.

\vspace{-0.3em}
\subsubsection{Analysis of $\lambda_1$ and $k$ for semantic regularization}
In our SP$^2$OT algorithm, the $\lambda_1$ affects the semantic regulation strength. As shown in Table \ref{tab:lambda1}, "500", "1000", "2000" is the value of $\lambda_1^0$, which is the initial value of $\lambda_1$ and decays with the training process as in Eq.\ref{eq:lambda1}. "Fix" donates $\lambda_1$ is fixed to 1000 during the training process. We conduct these experiments with $k$=20. we observed that a larger $\lambda_1$ is preferred for CIFAR100 and iNature500, while a smaller one is preferred for ImgNet-R. We suggest to set $\lambda_1$ as 1000, which achieves satisfactory results on all datasets.

Additionally, we can observe that when $\lambda_1$ is fixed at 1000, the performance of the model on CIFAR100 and ImgNet-R shows little improvement, while there is a sizeable decline on iNature500. This is because the k-NNG constructed on these two datasets is clean and reliable. In contrast, iNature500 has a larger number of categories, resulting in more noise within the k-NNG. Therefore, not applying weight decay would constrain the model's learning in this scenario, showing the necessity of semantic regularization decline.

The parameter $k$ is the number of nearest neighbors selected in the dataset for each sample in k-NNG. We conduct experiments with $k$=10, 20, 50 as shown in Table \ref{tab:topk}. In these experiments, we set $\lambda_1^0$=1000. Empirically, with a small $k$, the semantic regularization is too weak to guide the pseudo-label generation, while with a large $k$, the semantic regularization will introduce large noise from the initial feature space and hurt the performance. From the results, we can see that $k$=20 is a suitable choice for all three datasets and is the default setting in our experiments. Furthermore, when $k=20$, we evaluate the quality of the adjacency matrix $\mathbf{A}$ in Tab.~\ref{tab:am_acc}. An entry $\mathbf{A}_{ij} > 0$ is considered accurate if sample $i$ and $j$ belong to the same class. The results demonstrate the high quality of $\mathbf{A}$, which is more accurate than a random adjacency matrix.

\begin{table}[!t]
\caption{Analysis of $\lambda_1$ on different datasets.}

\label{tab:lambda1}
\centering
\resizebox{0.4\textwidth}{!}{
\begin{tabular}{c|ccc|c}
\toprule
$\lambda_1$  & 500  & 1000  & 2000 & Fix(1000)   \\ \midrule
CIFAR100     & 36.0 & 39.1  & 37.8 & 39.4  \\ 
ImgNet-R     & 28.7 & 27.1  & 27.2 & 27.8 \\
iNature500   & 31.9 & 34.1  & 34.0 & 32.3 \\
\bottomrule
\end{tabular}}
\end{table}

\begin{table}[!t]
    \caption{Analysis of $k$ for k-NNG on different datasets.}
    
    \label{tab:topk}
    \centering
    \resizebox{0.3\textwidth}{!}{
    \begin{tabular}{c|ccc}
    \toprule
    $k$  & 10  & 20 & 50  \\ \midrule
    CIFAR100     & 38.6 & 39.1 & 38.4  \\ 
    ImgNet-R     & 28.8 & 27.1 & 28.8 \\
    iNature500   & 32.2 & 34.1 & 32.6 \\
    \bottomrule
\end{tabular}}
\end{table}

\begin{table}[!t]
    \caption{The quality of the adjacency matrix $\mathbf{A}$. Values in parentheses is the accuracy of a random adjacency matrix.}
    
    \label{tab:am_acc}
    \centering
    \resizebox{0.48\textwidth}{!}{
    \begin{tabular}{c|cccc}
    \toprule
    $k=20$  & CIFAR100  & ImgNet-R & iNature100 & iNature500 \\ \midrule
    Acc (\%)     & 67.0 (2.5) & 30.1 (0.7) &  73.3 (1.2) & 55.4 (5.8)   \\ \bottomrule
\end{tabular}}
\end{table}

\vspace{-0.3em}
\subsubsection{Analysis of the kernel function}
In our method, we use the Gaussian kernel. To validate the effect of the different kernel functions, we further conduct experiments with the Cosine kernel, which is:
\begin{equation}
    \mathit{K_{cos}}(z_i, z_j) = \frac{ z_i \cdot z_j }{ || z_i || || z_j || }
\end{equation}
The corresponding results are detailed in Table \ref{tab:kernel}. In our experiments, we set $\lambda_1^0=1000$ and $k=20$. It is noteworthy that the Gaussian kernel outperforms the Cosine kernel on CIFAR100 and iNature500, while the Cosine kernel exhibits slightly better performance on ImgNet-R. These findings underscore the general applicability of our method across different kernels, showcasing its versatility and effectiveness in diverse experimental settings.
\begin{table}[!t]
    \caption{Analysis of the kernel function.}
    \label{tab:kernel}
    \centering
    \resizebox{0.3\textwidth}{!}{
    \begin{tabular}{c|cc}
    \toprule
    $\mathit{K}(\cdot, \cdot)$  & gaussian  & cosine  \\ \midrule
    CIFAR100     & 39.1 & 37.9  \\ 
    ImgNet-R     & 27.1 & 28.3 \\
    iNature500   & 34.1 & 33.5 \\
    \bottomrule
\end{tabular}}
\end{table}

\vspace{-0.3em}
\subsection{Limitation}
While our method demonstrates effective improvements, there are several avenues for exploration in future research. Firstly, our MM-based solver for SP$^2$OT requires iteration to optimize, necessitating a more efficient algorithm. Secondly, the curriculum strategy employed in our method follows a predetermined incremental approach, which is a suboptimal solution; therefore, a more adaptive strategy is warranted.









%% file: sec/conclu.tex
\section{Conclusions}
In this paper, we introduce a more practical problem known as "deep imbalanced clustering", which is designed to learn representations and semantic clusters from unlabeled imbalanced data, aiming to bridge the gap between practical scenarios and existing research. In response to this challenging problem, we present a novel progressive pseudo-label (PL)-based learning framework. This framework formulates the pseudo-label generation as a semantic regularized progressive partial optimal transport (SP$^2$OT) algorithm. The SP$^2$OT algorithm is instrumental in generating high-quality pseudo-labels by concurrently considering imbalanced cluster distribution, high-confidence sample selection, and semantic relations. This comprehensive approach significantly enhances model learning and clustering performance, offering a promising solution to the complexities inherent in deep imbalanced clustering scenarios.

To address the novel SP$^2$OT algorithm, we utilize the PMD algorithm. Initially, we derive the gradient for SP$^2$OT and perform projection through a Progressive Partial Optimal Transport (P$^2$OT) problem. To enhance the efficiency of solving P$^2$OT, we introduce a virtual cluster and incorporate a weighted $KL$ constraint. Subsequently, by imposing specific constraints, we transform P$^2$OT into an unbalanced optimal transport problem, amenable to an efficient solution through a scaling algorithm. Consequently, the SP$^2$OT is approximately solved through an iterative process that alternates between these two steps. To demonstrate the effectiveness of our approach, we establish a new benchmark comprising a human-curated long-tailed CIFAR100 dataset, challenging ImageNet-R datasets, and several large-scale fine-grained iNature datasets. Through extensive experiments on these datasets, we validate the superiority of our proposed method.

%% file: sec/supply.tex
\section*{Supplementary Material}
\renewcommand{\thefigure}{S\arabic{figure}}
\renewcommand{\thetable}{S\arabic{table}}
\renewcommand{\theequation}{S\arabic{equation}}

\renewcommand{\thelemma}{S\arabic{lemma}}




\section{Efficient Scaling Algorithm for solving Optimal Transport} \label{appendix:scaling_algorithm}
In this section, we detail how to solve the optimal transport by an efficient scaling algorithm. Let's recall the definition of optimal transport, given two probability vectors $\bm\mu\in\mathbb{R}^{m\times 1},\bm\nu\in\mathbb{R}^{n\times 1}$, as well as a cost matrix $\mathbf{C} \in \mathbb{R}^{m\times n}_+$ defined on joint space, the objective function which OT minimizes is as follows:
\begin{equation}\label{eq:general_form_appendix}
\min_{\mathbf Q \in \mathbb{R}^{m\times n}_+}\langle\mathbf{Q},\mathbf C\rangle_F + F_1(\mathbf{Q}\mathbf{1}_{n}, \bm\mu) + F_2(\mathbf{Q}^{\top} \mathbf1_m, \bm\nu)
\end{equation}
where $\mathbf Q \in \mathbb{R}^{m\times n}_+$ is the transportation plan, $\langle,\rangle_F$ denotes the Frobenius product, $F_1,F_2$ are constraints on the marginal distribution of $\mathbf Q$, which are convex, lower semicontinuous and lower bounded functions, and $\bm 1_n \in \mathbb{R}^{n\times 1}, \bm 1_m \in \mathbb{R}^{m\times 1}$ are all ones vector. To solve it efficiently, motivated by \cite{cuturi2013sinkhorn}, we first introduce an entropy constraint, $-\epsilon \mathcal{H}(\mathbf{Q})$. Therefore, the first term of Eq.(\ref{eq:general_form_appendix}) is as follows:  
\begin{align}
    ⟨\mathbf{Q},\mathbf C⟩_F - \epsilon \mathcal{H}(\mathbf{Q}) &= \epsilon <\mathbf{Q}, \mathbf{C}/\epsilon + \log \mathbf{Q}>_F \\
    &= \epsilon <\mathbf{Q}, \log \frac{\mathbf{Q}}{\exp(-\mathbf{C}/\epsilon)}>_F \\
    &=\epsilon KL (\mathbf{Q}, \exp(-\mathbf{C}/\epsilon)),
\end{align}
The entropic optimal transport can be reformulated as follows:
\begin{align}\label{eq:general_form_entropic_appendix}
\min_{\mathbf Q \in \mathbb{R}^{m\times n}_+}\epsilon KL (\mathbf{Q}, \exp(-\mathbf{C}/\epsilon)) + F_1(\mathbf{Q}\mathbf{1}_{n}, \bm\mu) + F_2(\mathbf{Q}^{\top} \mathbf1_m, \bm\nu)
\end{align}
Then, this problem can be approximately solved by an efficient scaling Alg.\ref{alg:saot}, where the proximal operator is as follows:
\begin{equation}\label{eq:prox}
     \text{prox}_{F/\epsilon}^{KL}(\mathbf{z}, \bm\mu) = \text{argmin}_{\mathbf{x}\in \mathbb{R}_{+}^n} F(\mathbf{x}, \bm\mu) + \epsilon KL(\mathbf{x},\mathbf{z})
\end{equation}

Intuitively, it is the iterative projection on affine subspaces for the KL divergence.
We refer readers to \cite{chizat2018scaling} for more derivation. Consequently, for OT problems with any proper constraint, if we can transform it into the form of Eq.(\ref{eq:general_form_appendix}) and derive the corresponding proximal operator, we can solve it with Alg.\ref{alg:saot} efficiently.









\SetKwComment{Comment}{//}{}
\begin{algorithm}[h]
\caption{Scaling Algorithm for Optimal Transport}\label{alg:saot}
\SetAlgoLined
\footnotesize
\DontPrintSemicolon
\KwIn{Cost matrix $\mathbf C$, $\epsilon,m,n,\bm\mu,\bm\nu$}    

$\mathbf{M}= \exp(-\mathbf{C}/\epsilon)$

$\mathbf{b} \leftarrow \mathbf{1}_{n} $

\While{$\bm b$ not converge}{

$\mathbf{a} \leftarrow \text{prox}_{F_1/\epsilon}^{KL}(\mathbf{M}  \mathbf{b}, \bm\mu)/(\mathbf{M}  \mathbf{b})$

$\mathbf{b} \leftarrow \text{prox}_{F_2/\epsilon}^{KL}(\mathbf{M}^\top  \mathbf{a}, \bm\nu)/(\mathbf{M}^\top  \mathbf{a})$
}

\KwRet $\text{diag}(\mathbf{a}) \mathbf{M} \text{diag}(\mathbf{b})$;

\end{algorithm}

\section{Majorization-Minimization for SP$^2$OT}
In this section, we propose a Majorization-Minimization (MM) algorithm to solve the SP$^2$OT problem when $\mathbf{A}$ is a positive semi-definite matrix. First, we outline the general framework of the MM algorithm. Next, we demonstrate that SP$^2$OT is a concave function. Finally, we present the resulting optimization procedure.

\subsection{Majorization-Minimization}
Majorization-Minimization (MM)~\cite{sun2016majorization} is a method for addressing non-convex optimization problems. This method exploits the inherent convexity within the function to find its minima through iterative processes. Specifically, let $f(\mathbf{Q})$ be a non-convex function to be minimized, and $\mathbf{Q}_t$ denote the solution at the $t$-th iteration. We first construct a convex and easy-to-solve surrogate function $g(\mathbf{Q}| \mathbf{Q}_t)$ that Majorizations the objective function, which satisfies:
\begin{align}\label{eq:mm1}
    g(\mathbf{Q}_t| \mathbf{Q}_t) = f(\mathbf{Q}_t)\\
    g(\mathbf{Q}| \mathbf{Q}_t) \geq f(\mathbf{Q}_t). \label{eq:mm2}
\end{align}
Then, we can minimize $g(\mathbf{Q}| \mathbf{Q}_t)$ to obtain $\mathbf{Q}_{t+1}$, which is guaranteed to be a better solution than $\mathbf{Q}_t$. By repeating this process, we can obtain a sequence of solutions $\{\mathbf{Q}_t\}$, which converges to a local minimum of $f(\mathbf{Q})$. 

Therefore, the key to minimizing the non-convex $f(\mathbf{Q})$ lies in identifying a suitable surrogate function $g(\mathbf{Q}| \mathbf{Q}_t)$. When dealing with a concave and differentiable function $f(\mathbf{Q})$, a viable approach to constructing such a surrogate function is to leverage the Taylor expansion of $f(\mathbf{Q})$, represented as:
\begin{align}
    f(\mathbf{Q}) \leq  f(\mathbf{Q}_t) + \nabla f(\mathbf{Q}_t) (\mathbf{Q} - \mathbf{Q}_t) = g(\mathbf{Q}| \mathbf{Q}_t)
\end{align}

\subsection{Proof of Concave}\label{appendix:concave}
\begin{prop}\label{prop:concave}
For the function:
\begin{equation}
    f(\mathbf{Q}) = \langle\mathbf{Q},-\log \mathbf P\rangle_F -  \lambda_1 \langle \mathbf{A}, \mathbf{Q}\mathbf{Q^\top} \rangle_F,
\end{equation}
where $\mathbf{Q}, \mathbf{A}, \mathbf{P}$ is some matrix, $\lambda_1 \geq 0$.\\
$f(\mathbf{Q})$ is a concave function on the feasible set $\mathbb{R}^{N\times K}_+$ if and only if $\mathbf{A}$ is a positive semi-definite matrix.\\

We know the gradient and hessian matrix of $f(\mathbf{Q})$ is as follows:
\begin{align}
    \nabla f(\mathbf{Q}) = -\log \mathbf P - \lambda_1 (\mathbf{A} + \mathbf{A}^\top)\mathbf{Q}\\
    \nabla^2 f(\mathbf{Q}) = - \lambda_1 (I \otimes (\mathbf{A} + \mathbf{A}^\top)), 
\end{align}
where $\nabla^2 f(\mathbf{Q}) \in R^{(N\times K)\times ((N\times K))}$. When $\mathbf{A}$ is positive semi-definite, then $\mathbf{A} + \mathbf{A}^\top$ is also positive semi-definite. Since $(\mathbf{A} + \mathbf{A}^\top)$ forms the diagonal of $I \otimes (\mathbf{A} + \mathbf{A}^\top)$, we know $I \otimes (\mathbf{A} + \mathbf{A}^\top)$ is also positive semi-definite. As $\lambda_1 \geq 0$, we have $\nabla^2 f(\mathbf{Q})$ is negative semi-definite. Therefore, $f(\mathbf{Q})$ is a concave function on the feasible set $\mathbb{R}^{N\times K}_+$.\\

\end{prop}







\subsection{Optimization for SP$^2$OT}
As shown in the previous section, $f(\mathbf{Q})$ is a concave function on the feasible set $\Pi$.
Then, at each iteration $t+1$, given the previously computed $\mathbf{Q}_t$, we construct a majorized objective function through Taylor expansion:
\begin{align}\label{eq:sppot_surrogate}
    g(\mathbf{Q}|\mathbf{Q}_t) = f(\mathbf{Q}_t) + \langle \nabla f(\mathbf{Q}_t), \mathbf{Q}-\mathbf{Q}_t \rangle_F \\ 
    \nabla f(\mathbf{Q}_t) = -\log \mathbf P -  \lambda_1 (\mathbf{A}+\mathbf{A^\top}) \mathbf{Q}_t.
\end{align}
It can be verified $f(\mathbf{Q}) \leq g(\mathbf{Q}|\mathbf{Q}_t)$ and $ f(\mathbf{Q}_t) = g(\mathbf{Q}_t | \mathbf{Q}_t) $, then
$g(\mathbf{Q}|\mathbf{Q}_t)$ is a valid majorized version of $f(\mathbf{Q})$.

Replacing the surrogate function with $f(\mathbf{Q})$ and ignoring the constant term, the updated SP$^2$OT is denoted as follows:
\begin{align}\label{eq:sppot_mm}
    \min_{\mathbf{Q} \in \Pi} \langle \mathbf{Q}, \nabla f(\mathbf{Q}_t)\rangle_F
    + \lambda_2 &KL(\mathbf{Q}^\top \bm1_N,\frac{\rho}{K}\bm 1_K) \\
    \text{s.t.}\quad \Pi = \{\mathbf{Q} \in \mathbb{R}^{N\times K}_+ | \mathbf{Q} \bm1_K &\leq\frac{1}{N}\bm1_N, \bm 1_N^\top\mathbf{Q} \bm 1_K=\rho\}.
\end{align}
The presented problem shares the same form as P$^2$OT, and it can be effectively addressed using the efficient scaling algorithm. Consequently, SP$^2$OT can be solved by the MM algorithm, which is the same as projected mirror descent.

\section{Proof of Proposition \ref{prop:equivalent}}\label{appendix:concave}
\begin{prop}\label{prop:equivalent}
If $\mathbf C = [f(\mathbf{Q}_t), \bm 0_N]$, and $\bm\lambda_{:K}=\lambda, \bm\lambda_{K+1} \rightarrow +\infty$, the optimal transport plan $\hat{\mathbf{Q}}^\star$ of Eq.(25) can be expressed as:
\begin{equation}
    \hat{\mathbf{Q}}^\star = [\mathbf{Q}^\star, \bm\xi^\star],
\end{equation}
where $\mathbf{Q}^\star$ is optimal transport plan of Eq.(17), and $\bm\xi^\star$ is the last column of $\hat{\mathbf{Q}}^\star$.
\end{prop}

Assume the optimal transportation plan of Eq.(25) is $\hat{\mathbf{Q}}^\star$, which can be decomposed as $[\bar{\mathbf{Q}}^\star, \bm\xi^\star]$, and the optimal transportation plan of Eq.(17) is $\mathbf{Q}^\star$. 

\textbf{Step 1.} We first prove $\bm\xi^{\star\top}\bm 1_N = 1-\rho$. To prove that, we expand the second $\hat{KL}$ term and rewrite it as follows:
\begin{align}
    &\hat{KL}(\hat{\mathbf{Q}}^{\star\top} \bm1_N, \bm\beta, \bm\lambda) \\
    &= \sum_{i=1}^{K} \bm\lambda_i  [ \bar{\mathbf{Q}}^{\star\top}\bm 1_N ]_i \log \frac{[\bar{\mathbf{Q}}^{\star\top}\bm 1_N]_i}{\bm\beta_i} + \bm\lambda_{K+1} \bm\xi^{\star\top}\bm 1_N \log \frac{\bm\xi^{\star\top}\bm 1_N}{1 - \rho} \\
    &= \lambda KL(\bar{\mathbf{Q}}^{\star\top}\bm1_N, \frac{\rho}{K} \bm 1_K) + \bm\lambda_{K+1} \bm\xi^{\star\top}\bm 1_N \log \frac{\bm\xi^{\star\top}\bm 1_N}{1 - \rho}
\end{align}

Due to $\bm\lambda_{K+1}\rightarrow +\infty$, if $\bm\xi^{\star\top}\bm 1_N$ is not equal to $1-\rho$, the cost of Eq.(25) will be $+\infty$. In such case, we can find a more optimal $\hat{\mathbf{Q}}^\dagger$, which satisfies $\bm\xi^{\dagger\top}\bm 1_N = 1-\rho$, and its cost is finite, contradicting to our assumption. Therefore, $\bm\xi^{\star\top}\bm 1_N = 1-\rho$, and $\hat{KL}(\hat{\mathbf{Q}}^{\star\top} \bm1_N, \bm\beta, \bm\lambda)=\lambda KL(\bar{\mathbf{Q}}^{\star\top}\bm1_N, \frac{\rho}{K} \bm 1_K)$.

\textbf{Step 2.} We then prove $\bar{\mathbf{Q}}^\star, \mathbf{Q}^\star$ are in the same constraint set. Due to 
\begin{equation}
    \hat{\mathbf{Q}}^\star \bm 1_{K+1} = [\bar{\mathbf{Q}}^\star, \bm\xi^\star] \bm 1_{K+1}
    = \bar{\mathbf{Q}}^\star \bm 1_K + \bm\xi^\star
    = \bm 1_N ,
\end{equation}
and $\bm\xi^\star \geq 0$, we know $\bar{\mathbf{Q}}^\star \bm 1_K = \bm 1_N - \bm\xi^\star \leq \bm 1_N$. Furthermore, 
\begin{equation}
    \bm 1_N^\top \hat{\mathbf{Q}}^\star \bm 1_{K+1} = \bm 1_N^\top (\bar{\mathbf{Q}}^\star \bm 1_K + \bm\xi^\star) = \bm 1_N^\top \bar{\mathbf{Q}}^\star \bm 1_K + \bm 1_N^\top\bm\xi^\star=1, 
\end{equation}
and $\bm\xi^{\star\top}\bm 1_N = \bm 1_N^\top \bm\xi^{\star} = 1-\rho$, we derive that 
\begin{equation}
    \bm 1_N^\top \bar{\mathbf{Q}}^\star \bm 1_K =1 -  \bm 1_N^\top\bm\xi^\star= 1- (1- \rho) = \rho.
\end{equation}

In summary, $\bar{\mathbf{Q}}^\star \in \{\bar{\mathbf{Q}}^\star \in \mathbb{R}^{N\times K} | \bar{\mathbf{Q}}^\star \bm1_K\leq\frac{1}{N}\bm1_N, \bm 1_N^\top\bar{\mathbf{Q}}^\star \bm 1_K=\rho\}$, which is the same $\Pi$ in Eq.(17).

\textbf{Step 3.} Finally, we prove $\bar{\mathbf{Q}}^\star = \mathbf{Q}^\star$. 
According to Proposition 2, $\mathbf C = [-\log \mathbf P, \bm 0_N]$. We plug it into  Eq.(25), and the cost of $\hat{\mathbf{Q}}^\star$ is as follows:
\begin{align}\label{eq:cost_optimal_2}
    &\langle\hat{\mathbf{Q}}^\star,\mathbf C\rangle_F + \hat{KL}(\hat{\mathbf{Q}}^{\star\top} \bm1_N, \bm\beta, \bm\lambda) - \epsilon \mathcal{H}(\hat{\mathbf{Q}}^\star) \\
    &= \langle[\bar{\mathbf{Q}}^\star, \bm\xi^\star], [-\log \mathbf P, \bm 0_N]\rangle_F + \lambda KL(\bar{\mathbf{Q}}^{\star\top}\bm1_N, \frac{\rho}{K} \bm 1_K) - \epsilon \mathcal{H}([\bar{\mathbf{Q}}^\star, \bm\xi^\star]) \\
    &= \langle\bar{\mathbf{Q}}^\star, -\log \mathbf P\rangle_F + \lambda KL(\bar{\mathbf{Q}}^{\star\top}\bm1_N, \frac{\rho}{K} \bm 1_K) - \epsilon \mathcal{H}(\bar{\mathbf{Q}}^\star) - \epsilon \mathcal{H}(\bm\xi^\star) \\
    &= C_1 - \epsilon \mathcal{H}(\bm\xi^\star)
\end{align}
And in Eq.(17), the cost for $\mathbf{Q}^\star$ is as follows:
\begin{equation}\label{eq:cost_optimal_1}
     \langle\mathbf{Q}^\star, -\log \mathbf P\rangle_F + \lambda KL(\mathbf{Q}^{\star\top}\bm1_N, \frac{\rho}{K} \bm 1_K) - \epsilon \mathcal{H}({\mathbf{Q}}^\star) = C_2
\end{equation}
From step 2, we know that $\bar{\mathbf{Q}}^\star, \mathbf{Q}^\star$ are in the same set. Consequently, the form of Eq.(\ref{eq:cost_optimal_2}) is the same as Eq.(\ref{eq:cost_optimal_1}), and in set $\Pi$, $\langle\mathbf{Q}, -\log \mathbf P\rangle_F + \lambda KL(\mathbf{Q}^\top\bm1_N, \frac{\rho}{K} \bm 1_K) - \epsilon \mathcal{H}(\mathbf{Q})$ is a convex function. 

If $C_1=C_2$, due to it is a convex function, $\bar{\mathbf{Q}}^\star = \mathbf{Q}^\star$. 

If $C_1 > C_2$, we can construct a transportation plan $\hat{\mathbf{Q}}^\dagger=[\mathbf{Q}^\star, \bm\xi^\star]$ for Eq.(25), which cost is $C_2$, achieving smaller cost than $\hat{\mathbf{Q}}^\star$. The results contradict the initial assumption that $\hat{\mathbf{Q}}^\star$ is the optimal transport plan. Therefore, $C_1$ can not be larger than $C_2$.

If $C_1 < C_2$, we can construct a transportation plan $\bar{\mathbf{Q}}^\star$ for Eq.(17), which cost is $C_1$, achieving smaller cost than $\mathbf{Q}^\star$. The results contradict the initial assumption that $\mathbf{Q}^\star$ is the optimal transport plan. Therefore, $C_1$ can not be smaller than $C_2$.

In conclusion, $C_1=C_2$ and $\bar{\mathbf{Q}}^\star = \mathbf{Q}^\star$, i.e. $\hat{\mathbf Q}^\star = [\mathbf Q^\star, \bm\xi^\star]$. We can derive $\mathbf Q^\star$ by omitting the last column of $\hat{\mathbf{Q}}^\star$. Therefore, Proposition 2 is proven.



\section{Proof of Proposition \ref{prop:solver}}\label{appendix:prop_solver}
\begin{prop}\label{prop:solver} (Restatement)
Adding a entropy regularization $-\epsilon\mathcal{H}(\hat{\mathbf{Q}})$ to Eq.(25), we can solve it by efficient scaling algorithm. We denote $\mathbf M = \exp (-\mathbf{C}/\epsilon), \bm f= \frac{\bm\lambda}{\bm\lambda + \epsilon},\bm\alpha=\frac{1}{N}\bm1_N$. The optimal $\hat{\mathbf{Q}}^\star$ is denoted as follows:
\begin{equation}
    \hat{\mathbf{Q}}^\star = {\emph{diag}}(\mathbf{a}) \mathbf{M} \emph{diag}(\mathbf{b}),
\end{equation}
where $\mathbf{a,b}$ are two scaling coefficient vectors and can be derived by the following recursion formula:
\begin{equation}
    \mathbf{a} \leftarrow \frac{\bm{\alpha}}{\mathbf{M} \mathbf{b}}, \quad \mathbf{b} \leftarrow (\frac{\bm{\beta}}{\mathbf{M}^\top \mathbf{a}})^{\circ\bm f},
\end{equation}
where $\circ$ denotes Hadamard power, i.e., element-wise power. The recursion will stop until $\bm b$ converges.
\end{prop}

From the preliminary section in the manuscript, we know the key to proving Proposition \ref{prop:solver} is to derive the proximal operator for $F_1,F_2$. Without losing generality, we rewrite the entropic version of Eq.(25) in a more simple and general form, which is detailed as follows:
\begin{align}\label{eq:re_curr_unbalanced_ot_2_general}
\min_{\hat{\mathbf{Q}} \in \Phi}\epsilon KL (\mathbf{Q}, \exp(&-\mathbf{C}/\epsilon)) + \hat{KL}(\mathbf{Q}^\top \bm1_N, \bm\beta, \bm\lambda) \\
\text{s.t.}\quad \Phi = \{\mathbf{Q} \in &\mathbb{R}^{N\times K}_+ | \mathbf{Q}\bm1_K=\bm\alpha\}, \quad 
\end{align}
In Eq.(\ref{eq:re_curr_unbalanced_ot_2_general}), the equality constraint means that $F_1$ is an indicator function, formulated as:
\begin{equation}
    F_1(\mathbf{x}, \bm\alpha) = \left\{
    \begin{aligned}
        &0, \quad \mathbf{x} = \bm\alpha \\
        &+\infty, \quad \text{otherwise}
    \end{aligned}
    \right.
\end{equation}
Therefore, we can plug in the above $F_1$ to Eq.(\ref{eq:prox}), and derive the proximal operator for $F_1$ is $\bm\alpha$. For the weighted $KL$ constraint, the proximal operator is as follows:
\begin{align}
    &\text{prox}_{F_2/\epsilon}^{KL}(\mathbf{z}, \bm\beta) \\
    &= \text{argmin}_{\mathbf{x}} \hat{KL}(\mathbf{x}, \bm\beta, \bm\lambda) + \epsilon KL(\mathbf{x},\mathbf{z}) \\
    &= \text{argmin}_{\mathbf{x}} \sum_i \bm\lambda_i \mathbf{x}_i \log \frac{\mathbf{x}_i}{\bm\beta_i} - \bm\lambda_i\mathbf{x}_i + \bm\beta_i + \epsilon \mathbf{x}_i \log  
 \frac{\mathbf{x}_i}{\mathbf{z}_i} - \epsilon\mathbf{x}_i + \mathbf{z}_i \\
    &= \text{argmin}_{\mathbf{x}} \sum_i \bm\lambda_i \mathbf{x}_i \log \frac{\mathbf{x}_i}{\bm\beta_i} - \bm\lambda_i\mathbf{x}_i + \epsilon \mathbf{x}_i \log \frac{\mathbf{x}_i}{\mathbf{z}_i} - \epsilon\mathbf{x}_i \\
    &= \text{argmin}_{\mathbf{x}} \sum_i (\bm\lambda_i + \epsilon) \mathbf{x}_i \log \mathbf{x}_i - (\bm\lambda_i \log \bm\beta_i + \bm\lambda_i +\epsilon\log\mathbf{z}_i + \epsilon)\mathbf{x}_i
\end{align}
The first line to the second is that we expand the unnormalized $KL$. The second line to the third is that we omit the variable unrelated to $\mathbf x$. The third line to the fourth line is that we reorganize the equation. Now, we consider a function as follows:
\begin{equation}
    g(x) = ax\log x - bx
\end{equation}
Its first derivative is denoted as:
\begin{equation}
    g'(x) = a + a\log x - b
\end{equation}
Therefore, $\text{argmin}_x g(x) = \exp (\frac{b-a}{a})$. According to this conclusion, we know that:
\begin{equation}
    \mathbf{x}_i = \exp (\frac{\bm\lambda_i \log \bm\beta_i +\epsilon\log\mathbf{z}_i}{\bm\lambda_i + \epsilon}) = \bm\beta_i^{\frac{\bm\lambda_i}{\bm\lambda_i + \epsilon}}\mathbf{z}_i^{\frac{\epsilon}{\bm\lambda_i + \epsilon}}
\end{equation}
And the vectorized version is as follows::
\begin{equation}
    \mathbf{x} = \bm\beta ^{\circ \bm{f}}\mathbf{z}^{\circ (1 - \bm{f})}, \quad \bm f = \frac{\bm\lambda}{\bm\lambda + \epsilon}.
\end{equation}

Finally, we plug in the two proximal operators into Alg.\ref{alg:saot}. The iteration of $\mathbf a, \mathbf b$ is as follows:
\begin{align}
    \mathbf{a} &\leftarrow \text{prox}_{F_1/\epsilon}^{KL}(\mathbf{M}  \mathbf{b}, \bm\alpha)/(\mathbf{M}  \mathbf{b}) = \frac{\bm{\alpha}}{\mathbf{M}\cdot \mathbf{b}} \\
    \mathbf{b} &\leftarrow \text{prox}_{F_2/\epsilon}^{KL}(\mathbf{M}^\top  \mathbf{a}, \bm\beta)/(\mathbf{M}^\top  \mathbf{a}) \\
    &= \bm\beta ^{\circ \bm{f}}(\mathbf{M}^\top  \mathbf{a})^{\circ (1 - \bm{f})} / (\mathbf{M}^\top  \mathbf{a}) \\
    &= (\frac{\bm\beta}{\mathbf{M}^\top  \mathbf{a}})^{\circ \bm f}
\end{align}

Consequently, Proposition \ref{prop:solver} is proved.

\section{More discussion on uniform constraint}
In deep clustering, the uniform constraint is widely used to avoid degenerate solutions. Among them, KL constraint is a common feature in many clustering baselines, such as SCAN~\cite{vangansbeke2020scan}, PICA\cite{huang2020pica}, CC\cite{li2021contrastive}, and DivClust\cite{divclust2023}. These methods employ an additional entropy regularization loss term on cluster size to prevent degenerate solutions, which is equivalent to the KL constraint. Different from above, \cite{huang2022learning} achieves uniform constraints by maximizing the inter-clusters distance between prototypical representations. By contrast, our method incorporates KL constraint in pseudo-label generation, constituting a novel OT problem. As demonstrated in the paper, our approach significantly outperforms these baselines.

\section{Datasets Details}\label{appendix:dd}
\begin{figure*}
    \centering
    \begin{minipage}{0.45\textwidth}
    \centering
        \includegraphics[scale=0.35]{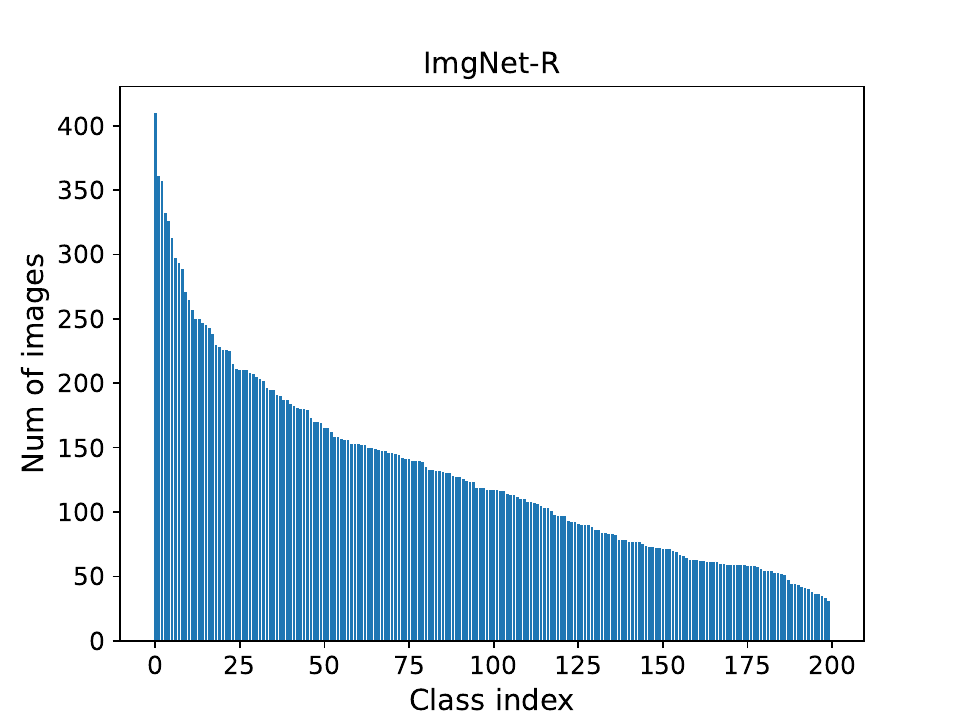}
    \end{minipage} %
    \begin{minipage}{0.45\textwidth}
    \centering
        \includegraphics[scale=0.35]{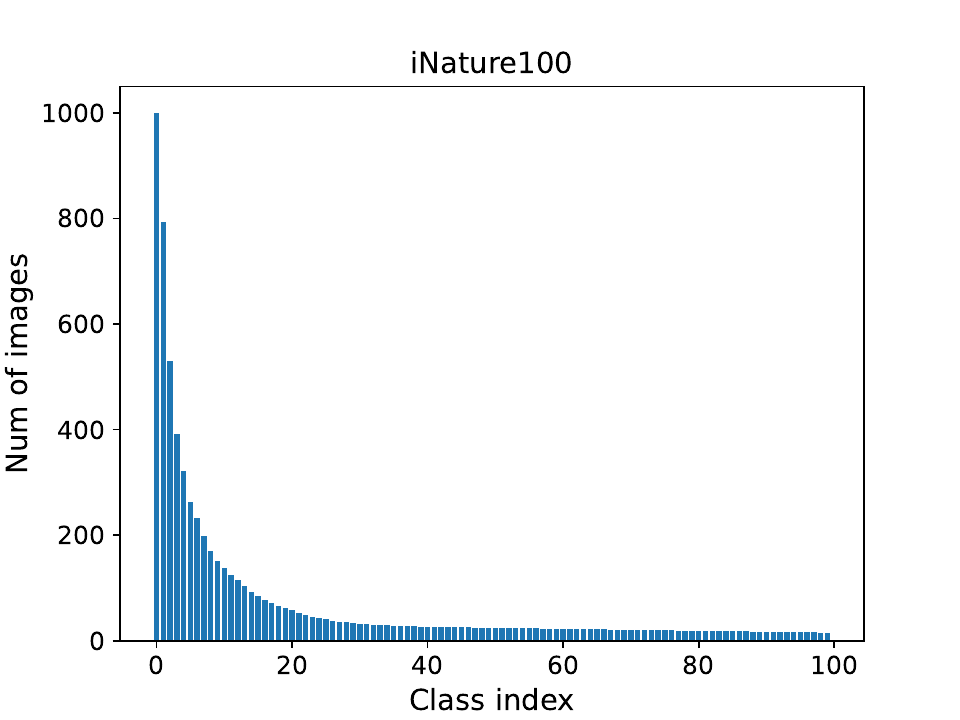}
    \end{minipage}
    \begin{minipage}{0.45\textwidth}
    \centering
        \includegraphics[scale=0.35]{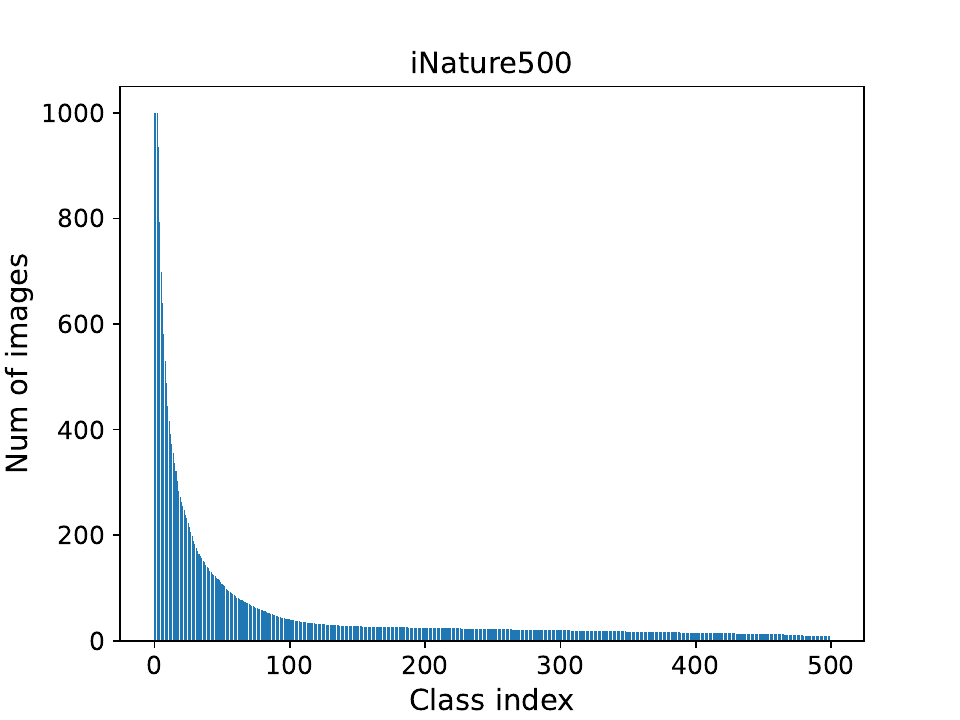}
    \end{minipage}%
        \begin{minipage}{0.45\textwidth}
        \centering
        \includegraphics[scale=0.35]{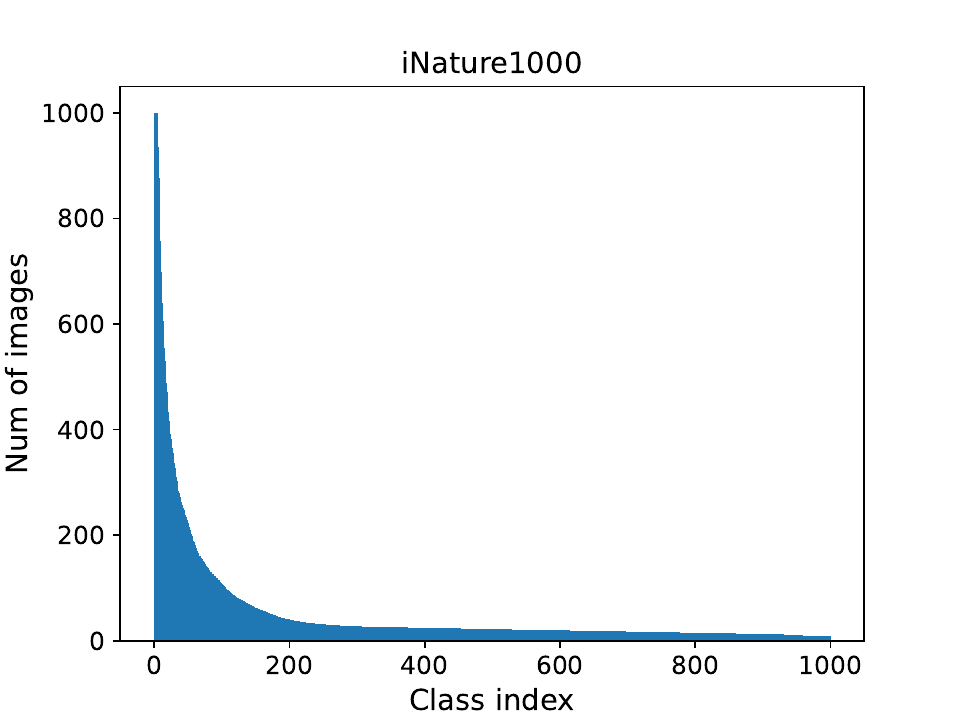}
    \end{minipage}
    \caption{The class distribution of different datasets.}
    \label{appendix:fig:class_dist}
    \vspace{-1em}
\end{figure*}

Fig.\ref{appendix:fig:class_dist} displays the class distribution of ImgNet-R and several training datasets from iNaturalist18. As evident from the figure, the class distribution in several iNature datasets is highly imbalanced. While ImgNet-R exhibits a lower degree of imbalance, its data distribution still presents a significant challenge for clustering tasks.


\section{More Implementation Details}\label{appendix:mid}
\subsection{Deep Clustering}
In our study, we employed various methods, all utilizing the same datasets and finetuning the last block of ViT-B16. All methods are trained for 50 epochs due to convergence observation.

For our P$^2$OT method, we use a batch size of 512. To stabilize the mini-batch P$^2$OT, We incorporate a memory buffer queue of size 5120 for history prediction after the first epoch to stabilize optimization. During training, we apply two transformations to the same sample and cross-utilize the two pseudo-labels as supervision, thereby encouraging the model to maintain eqivariance. We employ 2 clustering heads on the CIFAR100 dataset and 1 clustering head on others. The clustering heads are designed with the sample cluster number as ground truth. Since the sum of sample weights in the pseudo label $\mathbf Q$ is $\rho$, we adjust the loss on training sets for the clustering head and model selection as $\mathcal{L} / \rho$.

For IIC, we adopt their method (loss function) from \href{https://github.com/xu-ji/IIC}{GitHub}. They aim at maximizing the mutual information of class assignments between different transformations of the same sample. In all experiments, we use a batch size of 512. We follow their over-clustering strategy, i.e., using two heads, one matching the ground-truth cluster number and the other set at 200 for CIFAR100 and iNature100, 400 for ImgNet-R, and 700 for iNature500. We do not use over-clustering for iNature1000 due to resource and efficiency considerations. Models from the last epoch are selected.

For PICA, we adopt their method (loss function) from \href{https://github.com/Raymond-sci/PICA}{GitHub}. Their basic idea is to minimize cluster-wise Assignment Statistics Vector(ASV) cosine similarity. In our implementation, the batch size, over-clustering strategy, and model selection are the same as IIC above.

For SCAN, we adopt their code from \href{https://github.com/wvangansbeke/Unsupervised-Classification}{GitHub}. For clustering, their basic idea is to make use of the information of nearest neighbors from pretrained model embeddings. We use a batch size of 512. Following their implementation, we use 10 heads for CIFAR100, iNature100, iNature500, and ImgNet-R. We use 1 head for iNature1000 due to resource and efficiency considerations. While the authors select models based on loss on test sets, we argue that this strategy may not be suitable as test sets are unavailable during training. For the fine-tuning stage of SCAN, they use high-confident samples to obtain pseudo labels. Following their implementation, we use the confidence threshold of 0.99 on CIFAR. On other datasets, we use the confidence threshold of 0.7 because the model is unable to select confident samples if we use 0.99.

For CC, we adopt their code from \href{https://github.com/XLearning-SCU/2021-AAAI-CC}{GitHub}. They perform both instance- and cluster-level contrastive learning. Following their implementation, we use a batch size of 128, and the models of the last epochs are selected.

For DivClust, we adopt their code from \href{https://github.com/ManiadisG/DivClust}{GitHub}. Their basic idea is to enlarge the diversity of multiple clusterings. Following their implementation, we use a batch size of 256, with 20 clusterings for CIFAR100, ImgNet-R, iNature100, and iNature500. We use 2 clusterings for iNature1000 due to resource and efficiency considerations. The models of the last epoch along with the best heads are selected.

\begin{table*}[!h]
\centering
\caption{SPICE accuracy on different datasets.}
\vspace{-0.5em}
\label{tab:SPICE}
\resizebox{0.6\textwidth}{!}{
    \begin{tabular}{c|ccccc} \toprule
Dataset         & CIFAR100                                             & ImgNet-R              & iNature100            & iNature500           & iNature1000        \\ 
Imbalance ratio & 100                    & 13                    & 67                    & 111                  & 111               \\ \midrule
SPICE             &  6.74 &  2.58  &  10.29   &  2.15  &  0.74 \\

\bottomrule
    \end{tabular}}
\end{table*}

We also tried to implement SPICE from \href{https://github.com/niuchuangnn/SPICE}{GitHub} on imbalanced datasets. Given a pre-trained backbone, they employ a two-stage training process: initially training the clustering head while keeping the backbone fixed, and subsequently training both the backbone and clustering head jointly. For the first stage, they select an equal number of confident samples for each cluster and generate pseudo-labels to train the clustering head. However, on imbalanced datasets, this strategy can easily lead to degraded results as presented in Tab.\ref{tab:SPICE}. For example, for the medium and tail classes, the selected confident samples will easily include many samples from head classes, causing clustering head degradation. For the second stage, they utilize neighbor information to generate reliable pseudo-labels for further training. However, the degradation of the clustering head, which occurs as a result of the first stage's limitations, poses a hurdle in the selection of reliable samples for the second stage, thus preventing the training of the second stage. For comparison with SPICE on the balanced training set, please refer to Sec.\ref{sec:balance_train}.

\subsection{Representation Learning with long-tailed data}
In the literature, \cite{zhou2022contrastive,liu2021self,jiang2021self} aims to learn unbiased representation with long-tailed data. For a fair comparison, we adopt the pre-trained ViT-B16, finetune the last block of the ViT-B16 with their methods by 500 epochs, and perform K-means on representation space to evaluate the clustering ability. However, we exclude reporting results from \cite{liu2021self,jiang2021self} for the following reasons: 1) \cite{jiang2021self} prunes the ResNet, making it challenging to transfer to ViT; 2) \cite{liu2021self} has not yet open-sourced the code for kernel density estimation, which is crucial for its success.



\section{ARI metric}
Additionally, we report the results on the Adjusted Rand Index (ARI) metric, which is an instance-wise evaluation metric, on the balanced dataset. As shown in Tab.\ref{tab:test:ARI}, on the balanced test set, our method achieves comparable or superior performance to the previous SOTA. 
\section{NMI metric analysis}
The formula for the Normalized Mutual Information (NMI) is $\frac{2I(Y;C)}{H(Y)+H(C)}$, where Y represents the ground truth and C denotes the cluster label. Despite our method achieving superior clustering performance, as indicated by a higher $I(Y;C)$, it also results in a relatively uniform distribution, thereby increasing $H(C)$. This leads to only a modest improvement or decrease in the NMI metric.

\begin{table*}[!t]
\centering
\caption{ARI Comparison with SOTA methods on balanced test datasets.}
\vspace{-0.5em}
\label{tab:test:ARI}
\resizebox{0.7\textwidth}{!}{
    \begin{tabular}{c|cccccc} \toprule
Dataset         & CIFAR100                                             & ImgNet-R              & iNature100            & iNature500           & iNature1000        \\ 
Imbalance ratio & 100                    & 13                    & 67                    & 111                  & 111               \\ \midrule
IIC     &21.5$_{\pm 1.6}$             &11.6$_{\pm 0.9}$             &16.9$_{\pm 1.7}$             &7.7$_{\pm 0.4}$              &4.4$_{\pm 0.1}$              \\
PICA    &19.9$_{\pm 0.5}$             &5.2$_{\pm 0.0}$              &17.6$_{\pm 3.4}$             &6.4$_{\pm 0.3}$              &3.6$_{\pm 0.2}$              \\
SCAN    &\underline{28.2}$_{\pm 0.2}$ &12.0$_{\pm 0.5}$             &25.0$_{\pm 1.7}$             &17.0$_{\pm 0.3}$ &14.0$_{\pm 0.3}$ \\
SCAN*   &21.7$_{\pm 1.8}$             &\textbf{13.9}$_{\pm 0.1}$    &27.7$_{\pm 0.6}$ &11.1$_{\pm 0.6}$             &5.8$_{\pm 0.2}$              \\
CC      &19.1$_{\pm 0.7}$             &4.3$_{\pm 0.3}$              &15.1$_{\pm 0.4}$             &6.8$_{\pm 0.5}$              &4.3$_{\pm 0.1}$              \\
DivClust&21.3$_{\pm 0.6}$             &5.5$_{\pm 0.5}$              &19.2$_{\pm 1.1}$             &8.0$_{\pm 0.6}$              &4.5$_{\pm 0.3}$              \\
P$^2$OT &\textbf{28.8}$_{\pm 0.8}$    &\underline{13.4}$_{\pm 0.8}$ &\underline{29.7}$_{\pm 1.7}$    &\underline{18.9}$_{\pm 0.4}$    &\underline{14.2}$_{\pm 0.7}$    \\
SP$^2$OT &28.1$_{\pm 1.7}$    &13.1$_{\pm 0.1}$ &\textbf{31.9}$_{\pm 0.2}$    &\textbf{20.2}$_{\pm 0.8}$    &\textbf{15.8}$_{\pm 0.9}$    \\
 \bottomrule
    \end{tabular}}
\end{table*}
\begin{table*}[t]
\caption{Comparision with SPICE on balanced training set}
\label{tab:balanced_train}
\vspace{-0.5em}
\centering
\begin{tabular}{cc|c|c|c|c|ccc}
\toprule
 & \multirow{2}{*}{Method}& \multirow{2}{*}{Backbone} & Fine & Confidence   &  Time   &\multicolumn{3}{c}{CIFAR100}  \\ 
 &           &          &     Tune                           &    Ratio     &  Cost(h)   &ACC              & NMI             & ARI             \\ \midrule
 & SPICE$_s$ & ViT-B16  & False     & -                                     &  3.6   & 31.9             &57.7             & 22.5          \\ 
 & SPICE$_s$ & ViT-B16  & True      & -                                     &  6.7   & 60.9             &69.2             & 46.3          \\ 
 & SPICE     & ViT-B16  & False     & >0                                    &  3.6   & Fail             &Fail             & Fail           \\ 
 & SPICE     & ViT-B16  & True      & >0.6                                  &  6.7   & Fail             &Fail             & Fail           \\ 
 & SPICE     & ViT-B16  & True      & 0.5                                   &  21.6  & \textbf{68.0}    &\textbf{77.7}    & \textbf{54.8}  \\ 
 & Ours      & ViT-B16  & False     & -                                     &  2.9   & \underline{61.9} &\underline{72.1} & \underline{48.8}  \\ 
 \bottomrule
\end{tabular}
\end{table*}

\section{Comparison and analysis on balanced training dataset} \label{sec:balance_train}
To compare and analyze our method with baselines on balanced datasets, we conduct experiments using the balanced CIFAR100 dataset as demonstrated in Tab.\ref{tab:balanced_train}. Overall, we have achieved superior performance on imbalanced datasets, and on balanced datasets, we have achieved satisfactory results. However, in balanced settings, we believe that it is a bit unfair to compare our proposed method to methods developed in balanced scenarios, which usually take advantage of the uniform prior. 

Note that, as imagenet-r and iNaturelist2018 subsets are inherently imbalanced, we do not conduct experiments on them. For all methods, we employ the ImageNet-1K pre-trained ViT-B16 backbone. Fine-tune refers to the first training stage of SPICE, which trains the model on the target dataset using self-supervised learning, like MoCo. SPICE$_s$ corresponds to the second stage of training, which fixes the backbone and learns the clustering head by selecting top-K samples for each class, heavily relying on the uniform distribution assumption. SPICE refers to the final joint training stage, which selects an equal number of reliable samples for each class based on confidence ratio and applies FixMatch-like training. In SPICE, the confidence ratio varies across different datasets. Time cost denotes the accumulated time cost in hours on A100 GPUs.

The observed results in the table reveal several key findings:
1) Without self-supervised training on the target dataset, SPICE$_s$ yields inferior results;
2) SPICE tends to struggle without self-supervised training and demonstrates sensitivity to hyperparameters;
3) SPICE incurs a prolonged training time;
4) Our method achieves comparable results with SPICE$_s$ and inferior results than SPICE.

On imbalanced datasets, as demonstrated in Tab.\ref{tab:SPICE}, SPICE encounters difficulties in clustering tail classes due to the noisy top-K selection, resulting in a lack of reliable samples for these classes. In contrast, our method attains outstanding performance on imbalanced datasets without the need for further hyperparameter tuning.

In summary, the strength of SPICE lies in its ability to leverage the balanced prior and achieve superior results in balanced scenarios. However, SPICE exhibits weaknesses: 
1) It necessitates multi-stage training, taking 21.6 hours and 7.45 times our training cost;
2) It requires meticulous tuning of multiple hyperparameters for different scenarios and may fail without suitable hyperparameters, while our method circumvents the need for tedious hyperparameter tuning;
3) It achieves unsatisfactory results on imbalanced datasets due to its strong balanced assumption.

In contrast, the advantage of our method lies in its robustness to dataset distribution, delivering satisfactory results on balanced datasets and superior performance on imbalanced datasets.

\section{Further analysis of POT, UOT and SLA}\label{appendix:puot}
In this section, we provide a comprehensive explanation of the formulations of POT, UOT, and SLA~\cite{tai2021sinkhorn}, followed by an analysis of their limitations in the context of deep imbalanced scenarios. POT replaces the $KL$ constraint with an equality constraint, and its formulation is as follows:
\begin{align}\label{appendix:eq:pot}
\min_{\mathbf{Q} \in \Pi}\langle\mathbf{Q},-&\log \mathbf P\rangle_F \\
\text{s.t.}\quad \Pi = \{\mathbf{Q} \in \mathbb{R}^{N\times K}_+ &| \mathbf{Q} \bm1_K\leq\frac{1}{N}\bm1_N,  \notag \\ \mathbf{Q}^\top \bm1_N = \frac{\rho}{K} \bm1_K &, \bm 1_N^\top\mathbf{Q} \bm 1_K=\rho\}
\end{align}
This formulation overlooks the class imbalance distribution, thus making it hard to generate accurate pseudo-labels in imbalance scenarios. 

On the other hand, UOT eliminates the progressive $\rho$, and its corresponding formulation is denoted as follows:
\begin{align}\label{appendix:eq:uot}
\min_{\mathbf{Q} \in \Pi}\langle\mathbf{Q},-\log \mathbf P\rangle_F + \lambda KL(\mathbf{Q}^\top \bm1_N,\frac{1}{K}\bm 1_K) \\
\text{s.t.}\quad \Pi = \{\mathbf{Q} \in \mathbb{R}^{N\times K}_+ | \mathbf{Q} \bm1_K=\frac{1}{N}\bm1_N\}
\end{align}
UOT has the capability to generate imbalanced pseudo-labels, but it lacks the ability to effectively reduce the impact of noisy samples, which can adversely affect the model's learning process.

SLA~\cite{tai2021sinkhorn}, represented by Eq.(\ref{appendix:eq:sla}), relaxes the Eqality constraint using an upper bound $b$. However, during the early training stages when $\rho$ is smaller than the value in $b\bm 1_K$, SLA tends to assign all the mass to a single cluster, leading to a degenerate representation.
\begin{align}\label{appendix:eq:sla}
\min_{\mathbf{Q} \in \Pi}\langle\mathbf{Q},-&\log \mathbf P\rangle_F  \\
\text{s.t.}\quad \Pi = \{\mathbf{Q} \in \mathbb{R}^{N\times K}_+ | \mathbf{Q} \bm1_K\leq\frac{1}{N}\bm1_N, &\mathbf{Q}^\top \bm1_N \leq b \bm 1_K, \bm 1_N^\top\mathbf{Q} \bm 1_K=\rho\}
\end{align}

\section{Generalized Scaling Algorithm}\label{appendix:ea}
We provide the pseudo-code for the Generalized Scaling Algorithm (GSA) proposed by \cite{chizat2018scaling} in Alg.\ref{alg:gsa}. 

\begin{algorithm}[t!]
\caption{Generalized Scaling Algorithm for P$^2$OT}\label{alg:gsa}
\SetAlgoLined
\footnotesize
\DontPrintSemicolon
\KwIn{Cost matrix $-\log \mathbf P$, $\epsilon$, $\lambda$, $\rho$, $N,K$}    
$\mathbf{C} \leftarrow -\log \mathbf P$

$\bm \beta  \leftarrow \frac{\rho}{K} \bm 1_K^\top, \quad \bm{\alpha}  \leftarrow  \frac{1}{N}\mathbf{1}_N,  $

$\mathbf{b} \leftarrow \mathbf{1}_{K}, \quad s \leftarrow 1, \quad \mathbf{M} \leftarrow \exp(-\mathbf{C}/\epsilon), \quad \bm f \leftarrow \frac{\lambda}{\lambda + \epsilon}$

\While{$\bm b$ not converge}{

$\mathbf{a} \leftarrow min(\frac{\bm{\alpha}}{s\mathbf{M} \mathbf{b}}, 1)$

$\mathbf{b} \leftarrow (\frac{\bm{\beta}}{s\mathbf{M}^\top \mathbf{a}})^{\bm f}$

$s \leftarrow \frac{\rho}{\mathbf{a}^\top \mathbf{M} \mathbf{b}}$
}

\KwRet s$\text{diag}(\mathbf{a}) \mathbf{M} \text{diag}(\mathbf{b})$;

\end{algorithm}